\documentclass{article} 
\usepackage[]{iclr2027_conference,times}

\usepackage{amsmath,amsfonts,bm}

\def\eqref#1{equation~\ref{#1}}

\def\1{\bm{1}}

\DeclareMathAlphabet{\mathsfit}{\encodingdefault}{\sfdefault}{m}{sl}
\SetMathAlphabet{\mathsfit}{bold}{\encodingdefault}{\sfdefault}{bx}{n}

\usepackage{hyperref}
\usepackage{url}
\usepackage{graphicx}
\usepackage{amssymb}
\usepackage{adjustbox,capt-of}
\usepackage{soul}

\usepackage{xcolor}
\usepackage{colortbl}
\usepackage{booktabs}
\definecolor{highlightblue}{RGB}{234, 242, 253}
\definecolor{highlightpink}{RGB}{253, 237, 237}
\definecolor{azureblue}{rgb}{0.91, 0.94, 0.98}

\usepackage{enumitem}
\usepackage{wrapfig}
\usepackage{pifont}    
\usepackage{multirow}  
\newcommand{\cmark}{\ding{51}}  
\newcommand{\xmark}{\ding{55}}  

\title{Beyond Binary Preferences: Graded Preference Optimization for Limb-Motion Captioning}

\iclrfinalcopy
\makeatletter

\renewcommand{\maketitle}{%
  \par\begingroup
  \centering
  {\LARGE\bfseries \@title\par}
  \vspace{1em}
  {\normalsize\begin{tabular}{@{}c@{}}\@author\end{tabular}\par}
  \vspace{0.6em}
  \endgroup
  \thispagestyle{plain}%
}
\makeatother
\author{%
  \adjustbox{max width=\textwidth}{%
    \normalsize
    \textbf{Yanan Wang}$^{1,2,\ddagger}$ \quad 
    \textbf{Tingsong Li}$^{3,2,\ddagger}$ \quad 
    \textbf{Kaixun Jiang}$^{4,2,\ddagger}$ \quad 
    \textbf{Chenwei Xie}$^{2}$ \quad 
    \textbf{Chongyang Zhong}$^{2}$ \quad 
    \textbf{Zhaohe Liao}$^{5,2,\dagger}$%
  }\\[1mm]
  {\small $^1$Zhejiang University \quad $^2$Alibaba Token Hub, Alibaba Group}\\
  {\small $^3$University of Science and Technology of China \quad $^4$Fudan University \quad $^5$Shanghai Jiao Tong University}\\[1mm]
  {\small \texttt{\{mmwang@zju.edu.cn, zhaoheliao@sjtu.edu.cn\}}}\\[1mm]
  {\footnotesize $^{\ddagger}$Work done during internship at Alibaba Group. \quad
  $^{\dagger}$Corresponding author.}
}

\definecolor{myblue}{RGB}{173,216,230}

\renewenvironment{abstract}{%
  \par\medskip
  {\centering\large\bfseries \abstractname\par}%
  \smallskip
  \normalfont\normalsize
  \setlength{\parindent}{0pt}%
  \ignorespaces
}{%
  \par\medskip
}

\date{} 

\begin{document}
\maketitle

\noindent\begin{minipage}{\linewidth}
\centering
\includegraphics[width=0.95\linewidth]{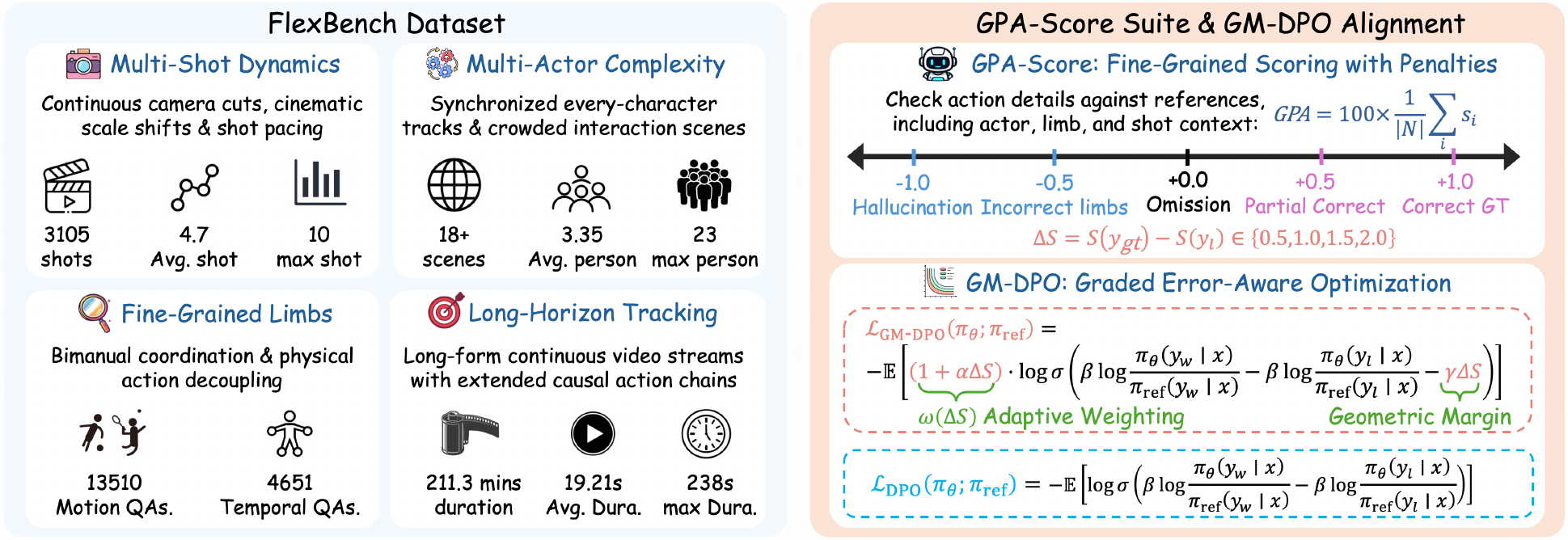}
\vspace{0.3em}
\captionof{figure}{Overview of our framework. FlexBench evaluates fine-grained limb motion for individual people across shots. GPA aggregates graded factual credit and explicit error penalties, while GM-DPO uses action-error severity to adjust preference margins and loss weights.}
\label{fig:teaser}

\end{minipage}\par\medskip

\begin{abstract}
Vision-Language Models (VLMs) can generate rich video
captions, yet often misidentify which person performs
an action or which limb is involved, particularly
across camera cuts.
Improving these details requires evaluation and
training that distinguish missing information from
incorrect assertions.
We introduce FlexBench, a benchmark spanning
3,105 shots and 18,161 evaluation queries, with
human-verified identities and systematic per-person
coverage of fine-grained limb actions and states.
Its reference-derived checklists support automated
assessment of complete captions in their person
and shot contexts.
Our Graded Physical Alignment score (GPA)
awards credit for correct content and deducts points
for incorrect or fabricated actions, making these
errors explicit in the aggregate score.
Building on this rubric, we propose
Graded Margin Direct Preference Optimization
(GM-DPO), which assigns stronger preference margins
and greater training weight to more severe action errors.
Across three VLM backbones, GM-DPO achieves the highest
substantive-action and GPA scores among the evaluated
preference objectives, improving GPA over DPO by
2.02--3.40 points.
On Qwen3-8B, it reduces the weighted hallucination rate
by 21.3\% relative to DPO.
These gains accompany sustained long-form output,
improved shot structure, and competitive performance
on three additional multimodal benchmarks.
\end{abstract}


\section{Introduction}
\label{sec:intro}

Video captioning turns visual observations into language, making video content accessible for understanding and downstream learning. Recent Vision-Language Models (VLMs) can produce rich captions covering scenes, people, and events~\citep{yang2025qwen3technicalreport,clark2026molmo2}, yet describing how an action unfolds remains challenging. Recognizing that someone opens a box, for example, does not establish which hand lifts the lid or how the other hand supports it. Across camera cuts and interacting people, these details must also remain attached to the correct person and shot. Our goal is to improve limb-motion fidelity within dense captions generated by general-purpose VLMs.

Progress toward this goal requires both targeted supervision and suitable evaluation. Supervised fine-tuning (SFT) teaches caption generation through reference imitation; subsequent preference optimization can further improve fidelity by contrasting better and worse captions~\citep{yuan2025tarsier2,lee2025vidchain}. Direct Preference Optimization (DPO)~\citep{rafailov2023direct} makes this practical with fixed preference pairs, without a separate reward model or online sampling during optimization. However, its binary preference labels do not explicitly distinguish omissions, limb confusions, and fabricated actions. This distinction matters when captions become training annotations: missing information leaves supervision incomplete, whereas false assertions introduce incorrect video--text associations. Evaluation presents a related challenge. Widely used multiple-choice question answering (MCQA) benchmarks~\citep{li2024mvbench,hong2025motionbench} measure answer selection, which does not directly reveal what a model would assert in an unrestricted caption. Although recent benchmarks directly evaluate motion captions~\citep{tu2025favorbench,lin2026kpmbench}, systematic limb-level assessment of every identifiable person across shots remains insufficiently addressed.

As shown in Figure~\ref{fig:teaser}, we address these needs through complementary advances in evaluation and training. We introduce FlexBench, a Fine-grained Limb-motion EXamination Benchmark, with human-verified identities and reference-derived checklists covering each person's actions in their corresponding shots. Coverage extends to stationary limbs and limb visibility, ensuring that evaluation encompasses limb states as well as movements. Our GPA combines fine-grained grades through a weighted average, awarding full or partial credit, assigning zero credit to omissions, and deducting points for incorrect or fabricated actions. Guided by the same grading rubric, GM-DPO assigns larger preference margins and loss weights to more severe action errors. This focuses preference learning on limb-motion fidelity within complete, richly detailed video captions.

Our main contributions are three-fold:
\begin{itemize}[leftmargin=2.5em,labelindent=0.5em,topsep=-4pt,itemsep=-1pt,parsep=0pt]
    \item \textbf{FlexBench:}
    A multi-shot benchmark spanning 3,105 shots and
    18,161 evaluation queries, with systematic
    per-person coverage of fine-grained limb motion,
    human-verified cross-shot identities, and
    checklists addressing actions, stationary limbs,
    and visibility.

    \item \textbf{GPA:}
    A graded, penalty-aware metric for automatically
    evaluating generated captions against detailed
    references, distinguishing incomplete coverage
    from incorrect motion claims.

    \item \textbf{GM-DPO:}
    An offline preference objective that incorporates
    action-error severity into margins and loss weights.
    Across three backbones, it achieves the best
    substantive-action and GPA scores among tested
    preference objectives while sustaining long-form
    output.
\end{itemize}

\vspace{-0.1em}

\section{Related Work}
\label{sec:related_work}

\subsection{From Video Understanding to Limb-Motion Captioning}

General-purpose VLMs increasingly capture not only
video events but also their temporal structure,
with benchmarks assessing long-video
comprehension~\citep{fu2024videomme,wu2024longvideobench}
and motion perception~\citep{hong2025motionbench}.
Limb-motion captioning requires finer granularity:
decomposing activities into constituent actions and
grounding each action in the correct person, limb,
and interaction.
KPM-Bench~\citep{lin2026kpmbench} advances fine-grained
motion captioning, accompanied by a training framework
combining multi-level motion representations with
SFT and GRPO.
Our work addresses the complementary challenge of
maintaining limb-level accuracy within dense captions
that cover every identifiable person across shots.
This scope includes stationary limbs and visibility,
enabling assessment of both action coverage and
unsupported motion claims.

\vspace{-0.2em}


\subsection{Preference Optimization for Video Captioning}

Preference learning complements reference imitation
with explicit comparisons between candidate outputs.
DPO~\citep{rafailov2023direct} provides an offline
objective for this supervision, while c-DPO,
IPO, and SimPO explore label smoothing, squared
preference objectives, and length-normalized rewards,
respectively~\citep{
mitchell2023cdpo,azar2023general,meng2024simpo}.
For video captioning,
Tarsier2~\citep{yuan2025tarsier2} applies DPO after
supervised training to improve detailed captions.
VidChain~\citep{lee2025vidchain} combines supervised
captioning and temporal grounding with metric-based
DPO, using task metrics to select preference pairs.
VideoComp~\citep{kim2025videocomp} instead trains
video--text matching models with a hierarchical
pairwise preference loss over increasingly disrupted
captions.
These studies motivate targeted preference supervision
for video understanding.
GM-DPO uses a limb-motion error rubric to determine
both the preference margin and the loss weight,
distinguishing incomplete action content from
incorrect or fabricated actions.

\subsection{Motion Benchmarks and Caption Evaluation}

General video benchmarks assess broad perceptual and temporal capabilities~\citep{li2024mvbench,fu2024videomme,wu2024longvideobench}. 
Motion-focused evaluations examine fine-grained dynamics via multiple-choice queries (MotionBench~\citep{hong2025motionbench}) and diverse QA-caption tasks (TempCompass~\citep{liu2024tempcompass}). 
Free-form generation additionally exposes the action
claims a model produces without candidate answers.
FAVOR-Bench~\citep{tu2025favorbench} evaluates motion
captions through LLM-assisted scoring and a separate
structured sequence-matching protocol.
Bridging these gaps, FlexBench emphasizes comprehensive, per-person and shot-specific limb assessments (covering stationary and hidden states). 
Our GPA metric resolves error severity via graded credits and penalties, supported by a temporal suite for consecutive-action coverage. 

\section{FlexBench: Fine-Grained Limb Physical Benchmarking}
\label{sec:flexbench}


FlexBench evaluates fine-grained limb-motion fidelity within complete multi-shot video captions. By pairing verified references with detailed checklists, it audits each person's active limbs and exact shot locations—enabling rigorous motion assessment without sacrificing descriptive richness.

\subsection{Data Collection and Quality Assurance Pipeline}
\label{sec:data_pipeline}

To prevent models from exploiting static background shortcuts, video sequences in FlexBench are curated from untrimmed cinematic scenes in AVA~\citep{gu2018ava} and CC-licensed YouTube streams, featuring domestic tasks, crafts, and complex tool manipulation.
As shown in Figure~\ref{fig:flexbench_overview}A, the collection spans varied interactions. Automated shot detection with manual boundary calibration yields 660 sequences comprising 3,105 shots, with a mean sequence duration of 19.21 seconds.

\begin{figure*}[t]
\centering
\includegraphics[width=0.97\linewidth, trim=0 2 0 0, clip]{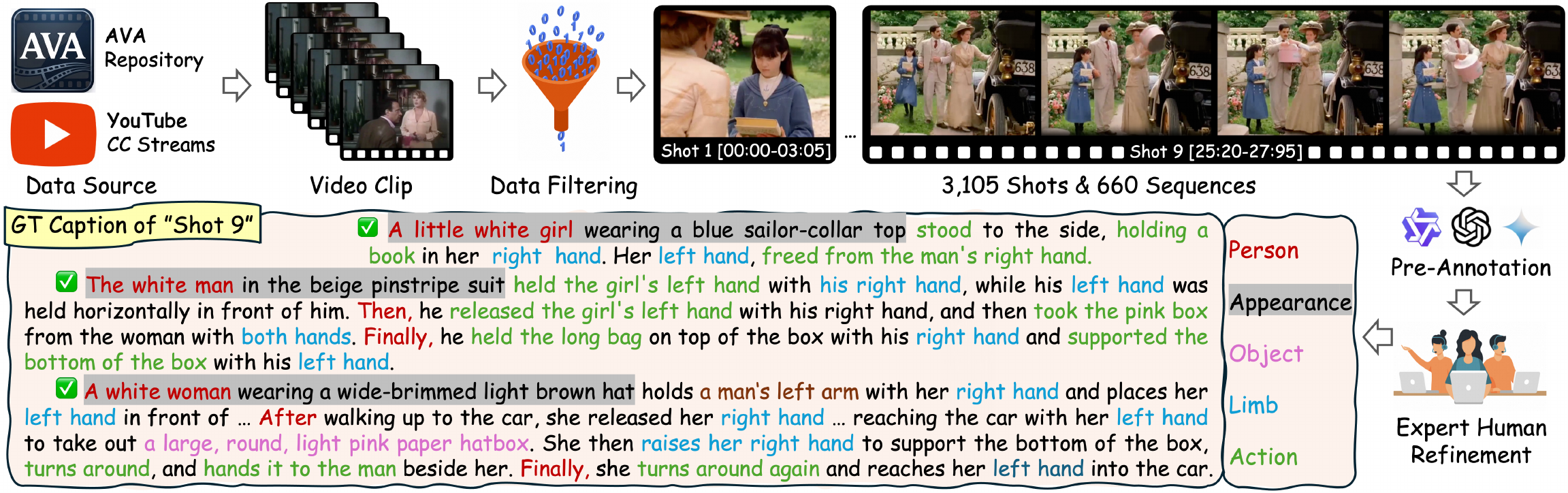}
\vspace{-1.0em}
\caption{FlexBench curation pipeline: video filtering, shot boundary calibration, VLM pre-labeling, and multi-stage expert refinement for granular limb-action modeling.}
\label{fig:data_pipeline}
\vspace{-0.6em}
\end{figure*}

As shown in Figure~\ref{fig:data_pipeline}, 18 trained annotators
refine VLM-generated captions to specify actor identities, interacted
objects, limb laterality, and action order. Cross-shot identities
are manually verified, with distinctive appearance cues used to
distinguish similar-looking people. Annotations undergo blind review
and consensus arbitration for ambiguous cases; details appear in
Appendix~\ref{sec:appendix_data_curation}.

\nocite{li2024mvbench,
        liu2024tempcompass,
        fu2024videomme,
        mangalam2023egoschema,
        wu2024longvideobench,
        hong2025motionbench,
        kim2025videocomp,
        lin2026kpmbench}

\subsection{Per-Person Coverage and Checklist Evaluation}
\label{sec:taxonomy}

\begin{figure*}[t]
\centering
\includegraphics[width=\linewidth, trim=0 2 0 0, clip]{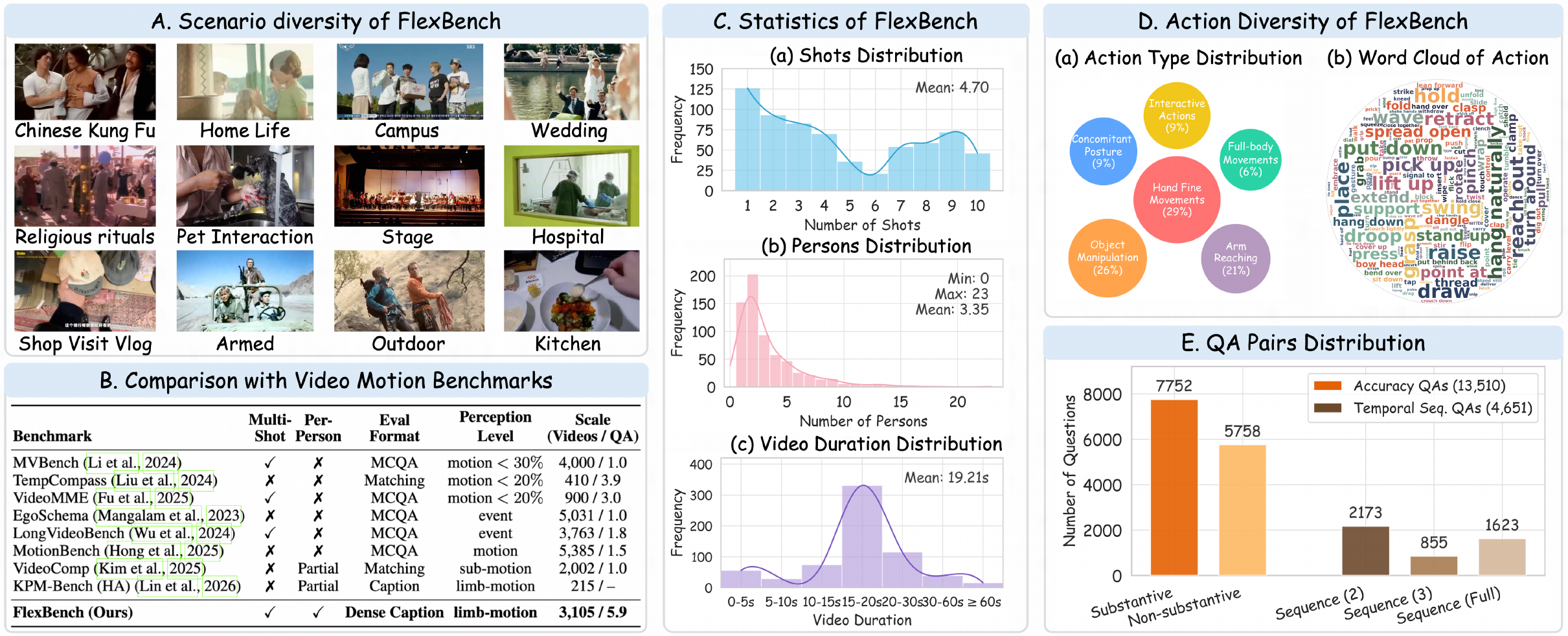}
\vspace{-1.7em}
\caption{Comprehensive overview of FlexBench: (A) Scenario coverage; (B) Dimensional comparison against existing video benchmarks; (C) Structural shot, actor, and duration distributions; (D) Kinematic action categories; (E) Action and temporal query counts.}
\label{fig:flexbench_overview}
\vspace{-0.8em}
\end{figure*}

\textbf{Coverage across people and shots.} Unlike conventional suites limited to isolated cuts or salient actors, FlexBench pairs multi-shot continuity with systematic, per-person limb annotation (Figure~\ref{fig:flexbench_overview}B). As shown in Figure~\ref{fig:flexbench_overview}C, sequences span 1–10 shots (mean 4.70) and 0–23 actors (mean 3.35). Reference captions structure each individual's actions by shot—resolving granular limb movements and object interactions while accounting for stationary or hidden limb states.

\textbf{Reference-derived factual checks.}
Human-verified captions are decomposed into positive checklist items covering actions, limb states, object interactions, and temporal order, each tied to a specific person and shot. During evaluation, an LLM judge scores the VLM-generated caption against both the checklist and ground-truth reference. 
Full credit strictly requires joint alignment across the action, person, and shot context—mismatched attributions receive no credit, while hallucinations and incorrect claims incur graded penalties. Human-evaluator agreement is analyzed in Section~\ref{sec:diagnostic_analysis}.

\subsection{Graded Evaluation}
\label{sec:gpa_score}

As shown in Figure~\ref{fig:flexbench_overview}E, FlexBench spans 18,161 queries. The Action Accuracy Suite comprises 7,752 substantive action queries and 5,758 non-substantive posture queries. The Temporal Sequence Suite contains 4,651 queries assessing consecutive actions in their shot contexts. Together, the two suites measure the fidelity of individual action claims and their joint coverage across sequences.

The Graded Physical Alignment score (GPA) assigns action-evaluation
items grades $s_i\in\{-1,-0.5,0,0.5,1\}$, distinguishing correct, 
partially specified actions, omissions, and contradictory or
fabricated actions. For category
$k\in\mathcal{K}=\{\mathrm{sub},\mathrm{non}\}$, let $N_k$ denote scored items we report
\vspace{-0.3em}
\begin{equation}
    \mathcal{S}_{\text{GPA}} = 100 \times \sum_{k \in \mathcal{K}} w_k \cdot \left( \frac{1}{|N_k|} \sum_{i \in N_k} s_i \right) = 100 \times \left( 0.6 \cdot \bar{S}_{\text{sub}} + 0.4 \cdot \bar{S}_{\text{non}} \right),
    \label{eq:gpa_score}
\end{equation}
\vspace{-0.1em}
Thus, grades use the $[-1,1]$ scale, whereas reported GPA scores lie
in $[-100,100]$. The clipped variant $\mathrm{GPA}_{\mathrm{clip}}$
replaces $s_i$ with $\max(0,s_i)$ in Equation~\ref{eq:gpa_score},
measuring credited content without negative penalties. We also
report the weighted hallucination rate
$\mathrm{WHR}=\mathrm{Ratio}_{-0.5}+2\mathrm{Ratio}_{-1}$,
where $\mathrm{Ratio}_v$ is the proportion of scored items assigned
grade $v$.

Temporal queries evaluate consecutive substantive actions within shot contexts. For example, a $\mathrm{Seq}_2$ item earns credit when both constituent actions are correctly represented. We report $\mathrm{TSA}_{\mathrm{w}}=0.2\mathrm{Seq}_2+0.3\mathrm{Seq}_3+0.5\mathrm{Seq}_{\mathrm{chain}}$, weighting two-action, three-action, and full-chain queries.

\section{GM-DPO: Graded Margin Preference Optimization}
\label{sec:methodology}

In this section, we formulate GM-DPO, an error-aware offline alignment objective to suppress fine-grained physical hallucinations in multimodal video understanding. GM-DPO addresses the core pathology of prevailing preference alignment algorithms: uniform treatment of non-preferred trajectories, obscuring the severe consequences of bodily chirality reversals and motion confabulations.

\subsection{Preliminaries and Limitations of Standard DPO}
\label{sec:dpo_prelim}

Given pairwise preferences $\mathcal{D} = \{(x, y_w, y_l)\}$ and a frozen reference policy $\pi_{\text{ref}}$, \citet{rafailov2023direct} optimizes a parameterized policy $\pi_\theta(y \mid x)$ under the Bradley-Terry model by minimizing:
\begin{equation}
\mathcal{L}_{\text{DPO}}(\pi_\theta; \pi_{\text{ref}}) = -\mathbb{E}_{(x, y_w, y_l) \sim \mathcal{D}} \left[ \log \sigma \left( h_\theta(x, y_w) - h_\theta(x, y_l) \right) \right],
\label{eq:standard_dpo}
\end{equation}
where $h_\theta(x, y) = \beta \log \frac{\pi_\theta(y \mid x)}{\pi_{\text{ref}}(y \mid x)}$ denotes the implicit reward. Differentiating Equation~\ref{eq:standard_dpo} with respect to $\theta$ yields the parameter gradient:
\begin{equation}
\left\{
\begin{aligned}
&\nabla_\theta \mathcal{L}_{\text{DPO}} = -\beta \cdot \mathbb{E}_{(x, y_w, y_l) \sim \mathcal{D}} \left[ \sigma \left( h_\theta(x, y_l) - h_\theta(x, y_w) \right) \cdot \Delta_\theta \log \pi(y_w, y_l) \right], \\
&\Delta_\theta \log \pi(y_w, y_l) \triangleq \nabla_\theta \log \pi_\theta(y_w \mid x) - \nabla_\theta \log \pi_\theta(y_l \mid x),
\end{aligned}
\right.
\label{eq:dpo_grad}
\end{equation}

Crucially, standard DPO assumes uniform binary preferences ($y_w \succ y_l$), scaling updates solely by prediction error $\sigma(h_\theta(x, y_l) - h_\theta(x, y_w))$ regardless of defect severity. Consequently, it applies identical gradient penalties to benign omissions and fatal chirality inversions (e.g., swapping hands), failing to penalize critical physical violations without over-correcting harmless variations.

\subsection{Perturbation Severity from the Grading Rubric}
\label{sec:graded_delta}

Each preference pair applies
one error category $e$ consistently to the targeted action
descriptions within a selected shot, retaining the remaining
caption. This restriction preserves the surrounding context
and is intended to avoid preference pairs that are easily
distinguished through errors distributed across multiple shots.
The original targeted content has
grade $1$, and the perturbation is assigned
$s_{\mathrm{pert}}(e)\in\{0.5,0,-0.5,-1\}$
under the action-grading rubric. We define
\begin{equation}
\Delta S
=1-s_{\mathrm{pert}}(e)
\in\{0.5,1,1.5,2\}.
\label{eq:delta_s}
\end{equation}
This label represents local perturbation severity, not the GPA
difference between complete captions. It uses the unscaled rubric,
without the factor of 100 used for benchmark reporting.
The preference objective below evaluates likelihoods over
complete captions.

\subsection{GM-DPO Objective and Optimization Dynamics}
\label{sec:gm_dpo_formulation}


Rather than enforcing uniform margins, GM-DPO governs policy updates through two coordinated mechanisms modulated by $\Delta S$: an internal dynamic geometric margin $\gamma \Delta S$ enforcing separation, and an external gradient weight $\omega(\Delta S) = 1 + \alpha \Delta S$ ($\alpha \ge 0, \gamma > 0$):
\begin{equation}
\mathcal{L}_{\text{GM-DPO}}(\pi_\theta; \pi_{\text{ref}}) = -\mathbb{E}_{(x, y_w, y_l) \sim \mathcal{D}} \left[ \omega(\Delta S) \log \sigma \left( h_\theta(x, y_w) - h_\theta(x, y_l) - \gamma \Delta S \right) \right].
\label{eq:gm_dpo_loss}
\end{equation}
Differentiating Equation~\ref{eq:gm_dpo_loss} with respect to model parameters $\theta$ yields the explicit gradient:
\vspace{-0.2em}
\begin{equation}
\nabla_\theta \mathcal{L}_{\text{GM-DPO}} = -\beta \cdot \mathbb{E}_{\mathcal{D}} \left[ \underbrace{\omega(\Delta S)}_{\text{Scaling}} \cdot \underbrace{\sigma \left( h_\theta(x, y_l) - h_\theta(x, y_w) + \gamma \Delta S \right)}_{\text{Margin-Shifted Residual Error}} \cdot \Delta_\theta \log \pi(y_w, y_l) \right].
\label{eq:gmdpo_grad}
\end{equation}
Equation~\ref{eq:gmdpo_grad} reveals the dual modulation mechanics of GM-DPO for vision-language alignment:

\textit{(1) Margin-Shifted Residual Error:} The offset $+\gamma \Delta S$ shifts the logistic saturation boundary. When $h_\theta(y_w) > h_\theta(y_l)$, standard DPO gradients rapidly vanish; in contrast, GM-DPO sustains gradients until clearing a severity-scaled margin, ensuring continuous optimization on subtle dynamics.

\textit{(2) Magnitude Scaling:} The multiplier $\omega(\Delta S)$ linearly magnifies parameter updates for severe physical flaws (up to $(1+2\alpha)\times$ for confabulations), penalizing chirality flips more aggressively than harmless omissions. This error-sensitive scaling functions as an adaptive step size, guiding the policy away from catastrophic hallucination regimes without destabilizing language representation.

\begin{figure*}[t]
\centering
\includegraphics[width=0.99\linewidth, trim=0 2 0 0, clip]{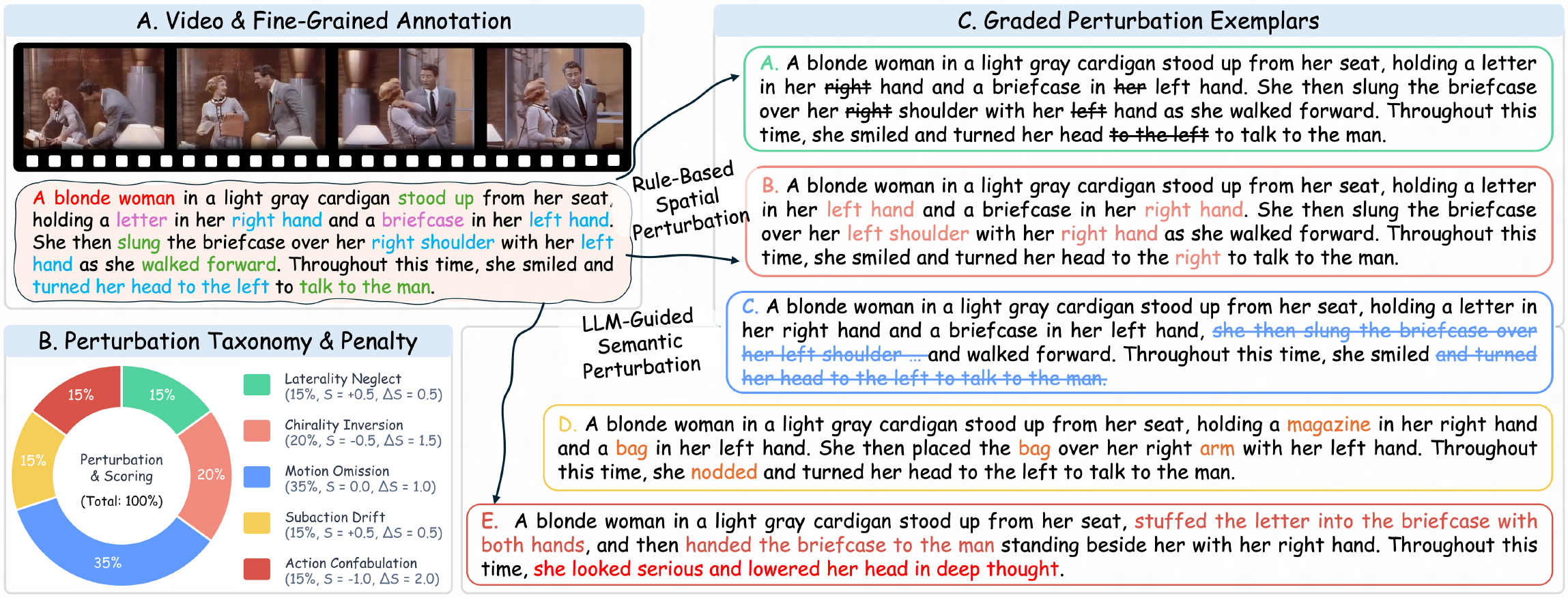}
\vspace{-0.9em}
\caption{The preference pair curation pipeline and perturbation spectrum. Ground-truth dense captions ($y_w, S=+1.0$) are perturbed via rule-based spatial editing and schema-constrained LLM rewriting to synthesize negative samples across five graded physical error regimes ($\Delta S \in [0.5, 2.0]$).}
\label{fig:pair_generation_pipeline}
\vspace{-0.6em}
\end{figure*}

\subsection{Preference Pair Construction}
\label{sec:pair_generation}

We construct 125k preference pairs from single- and multi-shot videos, using expert-verified captions as chosen responses. To prevent data contamination, the training pool is completely disjoint from FlexBench: it strictly excludes AVA footage and shares no source videos (e.g., films or YouTube streams) with the 660 evaluation sequences.

As illustrated in Figure~\ref{fig:pair_generation_pipeline},
rule-based editing produces Laterality Neglect ($\Delta S=0.5$)
and Chirality Inversion ($1.5$). Schema-constrained
Gemini-3.5-Flash rewriting generates Motion Omission ($1$),
Subaction Drift ($0.5$), and Action Confabulation ($2$).
Untargeted descriptions are retained, and rewrites are constrained
to preserve the original style. A double-blind study yielded
49.60\% human accuracy in distinguishing human-refined from
LLM-modified captions; generation and verification protocols are
provided in Appendix~\ref{sec:perturbation_validation}.

\section{Experiments}
\label{sec:experiments}

\vspace{-0.4em}
We evaluate whether GM-DPO improves limb-level action fidelity
within long-form, shot-structured video captions. Our experiments
first characterize existing models on FlexBench, then compare
various preference optimization methods, examine the
contributions of the proposed objective, and assess performance
on broader multimodal benchmarks.

\vspace{-0.4em}

\subsection{Experimental Setup}
\label{sec:exp_setup}

\textbf{Evaluated Model Suites.} We benchmark physical motion perception across three representative foundation tiers: (1) \textit{Proprietary Frontier APIs}: Gemini-3.1-Pro, Gemini-3.5-Flash~\citep{gemini31pro2026,gemini35flash2026}, Seed-2.1-Pro~\citep{bytedanceseed2026}, Kimi-k2.6~\citep{kimiteam2026kimik25visualagentic}, alongside Qwen-3.5-Omni ~\citep{qwenteam2026qwen35omnitechnicalreport} and Qwen-3.8-Max~\citep{qwenteam2026qwen38next}; (2) \textit{Open-Source Models}: Nemotron-3-Omni-30B~\citep{nvidia2025nemotron3nanoope}, Gemma-4-31B~\citep{gemmateam2026gemma4technicalreport}, Qwen3-32B~\citep{yang2025qwen3technicalreport}, and Qwen-3.8-27B~\citep{qwenteam2026qwen38next}. (3) \textit{Open-Weight Video Models (7B)}: Tarsier2-Recap-7B~\citep{yuan2025tarsier2} and VideoLLaMA2-7B~\citep{cheng2024videollama2}. To ensure comparability and consistency, all models are evaluated under identical prompts.

\textbf{Training Setup.}
We compare GM-DPO with DPO~\citep{rafailov2023direct},
c-DPO~\citep{mitchell2023cdpo},
IPO~\citep{azar2023general}, and
SimPO~\citep{meng2024simpo} on Qwen2.5-7B, Qwen3-8B,
and LLaVA-2-8B, with an additional SFT baseline on
Qwen3-8B.
All preference methods include a chosen-response NLL term,
$\mathcal{L}_{\mathrm{total}}
=\mathcal{L}_{\mathrm{pref}}+\lambda\mathcal{L}_{\mathrm{NLL}}$,
with $\lambda=1$.
Experiments are conducted on a cluster of 8$\times$NVIDIA A800 (80GB) GPUs using the AdamW optimizer, updating language backbone parameters while keeping visual encoders frozen.
For both SFT and preference tuning, runs are repeated across three random seeds with mean metrics reported.
Training configurations and baseline objectives appear
in Appendix~\ref{sec:appendix_training_details};
results with standard deviations are provided in
Appendix~\ref{sec:appendix_seed_robustness}.

\textbf{Evaluation Metrics.}
We report GPA, $\mathrm{TSA}_{\mathrm{w}}$, and their sub-metrics (averaged across GPT-4o, Gemini-3.5-Flash, and Qwen-3.8-Max) to quantify motion fidelity and temporal continuity. Concurrently, we record caption length ($\text{Len}$, in words) and shot structure match rate ($\text{SMR}$). 
``N/S'' denotes outputs failing to follow the requested multi-shot structure.
For these models, Motion Accuracy is evaluated without shot matching, while retaining checks on action, person, and limb correctness.

\subsection{Evaluating Existing Models on FlexBench}
\label{sec:zeroshot_profiling}

As shown in Table~\ref{tab:sota_profiling}, substantive-action scores are consistently lower than non-substantive scores across evaluated models: Qwen-3.8-Max reaches 33.50 versus 46.88, and Gemini-3.1-Pro reaches 17.80 versus 42.21. Fine-grained limb-motion captioning thus remains challenging even for models producing rich captions. Tarsier2-Recap-7B achieves a notable 22.08 GPA at just 127.9 average words, though evaluated under relaxed shot matching. The gap between $\mathrm{GPA}_{\mathrm{clip}}$ and GPA shows correct content often coexists with errors in dense captions: Qwen-3.8-Max and Seed-2.1-Pro incur deductions of 8.26 and 7.82 points, respectively.

Sequence-level scores reveal a related limitation.
Seed-2.1-Pro achieves the highest
$\mathrm{TSA}_{\mathrm{w}}$ at 46.30, while Qwen3-32B
leads the evaluated large open-weight models at 27.83.
These queries require consecutive actions to be
correctly represented in their corresponding shot
contexts, making their joint coverage more demanding.
Output structure alone does not resolve this difficulty:
Qwen-3.5-Omni achieves 99.34\% SMR but only 13.72
on substantive actions.
Conversely, the two evaluated 7B video models do not
produce the requested shot structure.
Together, these results motivate improving limb-motion
fidelity within complete, shot-organized captions.

\definecolor{azureblue}{rgb}{0.91, 0.94, 0.98}

\begin{table*}[t]
\vspace{-1.5em}
\caption{Zero-shot performance on FlexBench across SOTA VLMs. Best results are in \textbf{bold}.}
\label{tab:sota_profiling}
\vspace{-1.2em}
\begin{center}
\resizebox{1\linewidth}{!}{
\setlength{\tabcolsep}{5.2pt}
\begin{tabular}{l|cccc|cccc|cc}
\toprule
\multicolumn{1}{c|}{\bf Model} & \multicolumn{4}{c|}{\bf Motion Accuracy} & \multicolumn{4}{c|}{\bf Motion Sequence} & \multicolumn{2}{c}{\bf Output Statistics} \\
& \multicolumn{1}{c}{\bf Subs.} & \multicolumn{1}{c}{\bf Non-subs.} & \multicolumn{1}{c}{\cellcolor{azureblue}\bf GPA} & \multicolumn{1}{c|}{$\bf GPA_{\text{clip}}$} & \multicolumn{1}{c}{$\bf Seq_2$} & \multicolumn{1}{c}{$\bf Seq_3$} & \multicolumn{1}{c}{$\bf Seq_{\text{chain}}$} & \multicolumn{1}{c|}{\cellcolor{azureblue}$\bf TSA_{\text{w}}$} & \bf Len & \bf SMR (\%) \\
\midrule
\multicolumn{11}{l}{\textit{\textcolor{gray}{Proprietary Frontier APIs}}} \\
Gemini-3.1-Pro          & 17.80          & 42.21          & \cellcolor{azureblue}27.56          & 34.27          & 26.53          & 28.18          & 30.04          & \cellcolor{azureblue}28.78          & 855.7  & 50.27 \\
Gemini-3.5-Flash        & 16.70          & 43.61          & \cellcolor{azureblue}27.46          & 33.98          & 27.03          & 26.88          & 28.77          & \cellcolor{azureblue}27.86          & 709.4  & 56.22 \\
Seed-2.1-Pro            & 32.24          & 44.38          & \cellcolor{azureblue}37.10          & 44.92          & \textbf{47.00} & 44.14          & \textbf{47.31} & \cellcolor{azureblue}\textbf{46.30} & 782.3  & 68.31 \\
Kimi-k2.6               & 18.47          & 42.13          & \cellcolor{azureblue}27.93          & 36.04          & 30.35          & 30.26          & 32.94          & \cellcolor{azureblue}31.62          & 683.4  & 53.57 \\
Qwen-3.8-Max            & \textbf{33.50} & \textbf{46.88} & \cellcolor{azureblue}\textbf{38.85} & \textbf{47.11} & 45.24          & \textbf{45.44} & 46.70          & \cellcolor{azureblue}46.03          & 1233.3 & 99.18 \\
\midrule
\multicolumn{11}{l}{\textit{\textcolor{gray}{Open-Source Large Models}}} \\
Nemotron-3-Omni-30B     & 5.94           & 41.22          & \cellcolor{azureblue}20.05          & 25.10          & 14.92          & 16.24          & 15.22          & \cellcolor{azureblue}15.47          & 310.4  & 64.23 \\
Gemma-4-31B             & 7.63           & \textbf{48.37} & \cellcolor{azureblue}23.93          & 26.49          & 13.68          & 15.90          & 17.12          & \cellcolor{azureblue}16.07          & 424.9  & 78.98 \\
Qwen3-32B               & \textbf{14.20} & 39.91          & \cellcolor{azureblue}24.48          & \textbf{33.81} & \textbf{25.95} & \textbf{25.96} & \textbf{29.70} & \cellcolor{azureblue}\textbf{27.83} & 674.9  & 96.57 \\
Qwen-3.8-27B            & 13.47          & 44.53          & \cellcolor{azureblue}\textbf{25.89} & 31.66          & 22.76          & 23.04          & 25.28          & \cellcolor{azureblue}24.10          & 596.7  & \textbf{98.10} \\
\midrule
\multicolumn{11}{l}{\textit{\textcolor{gray}{Open-Weight Video Models (7B)}}} \\
VideoLLaMA2-7B          & 2.01             & 21.47            & \cellcolor{azureblue}9.79             & 13.89             & 6.84             & 4.62             & 9.68             & \cellcolor{azureblue}7.59             & 142.4     & N/S     \\
Tarsier2-Recap-7B       & \textbf{8.96}  & \textbf{41.75} & \cellcolor{azureblue}\textbf{22.08} & \textbf{24.98} & \textbf{9.99}           & \textbf{6.84}           & \textbf{12.17}          & \cellcolor{azureblue}\textbf{10.13}          & 127.9  & N/S    \\
\bottomrule
\end{tabular}
}
\end{center}
\vspace{-1.2em}
\end{table*}

\begin{table*}[t]
\caption{Main alignment results on FlexBench. Base models in \textit{\textcolor{gray}{gray italics}}; best aligned in \textbf{bold}.}
\label{tab:main_alignment_results}
\vspace{-1.2em}
\begin{center}
\resizebox{1\linewidth}{!}{
\begin{tabular}{l|cccc|cccc|cc}
\toprule
\multicolumn{1}{c|}{\bf Model Variant} & \multicolumn{4}{c|}{\bf Motion Accuracy} & \multicolumn{4}{c|}{\bf Motion Sequence} & \multicolumn{2}{c}{\bf Output Statistics} \\
& \multicolumn{1}{c}{\bf Subs.} & \multicolumn{1}{c}{\bf Non-subs.} & \multicolumn{1}{c}{\cellcolor{azureblue}\bf GPA} & \multicolumn{1}{c|}{$\bf GPA_{\text{clip}}$} & \multicolumn{1}{c}{$\bf Seq_2$} & \multicolumn{1}{c}{$\bf Seq_3$} & \multicolumn{1}{c}{$\bf Seq_{\text{chain}}$} & \multicolumn{1}{c|}{\cellcolor{azureblue}$\bf TSA_{\text{w}}$} & \multicolumn{1}{c}{\bf Len} & \multicolumn{1}{c}{\bf SMR (\%)} \\
\midrule
Qwen3-8B-SFT             & 6.92          & 42.31          & \cellcolor{azureblue}21.08          & 29.88          & 13.30          & 15.54          & 15.81          & \cellcolor{azureblue}15.23          & 1634          & 89.7          \\
\midrule
\textcolor{gray}{\textit{Qwen2.5-7B-Base}}   & \textcolor{gray}{\textit{3.02}}  & \textcolor{gray}{\textit{23.17}} & \cellcolor{azureblue}\textcolor{gray}{\textit{11.08}}  & \textcolor{gray}{\textit{14.26}} & \textcolor{gray}{\textit{8.86}}  & \textcolor{gray}{\textit{9.88}}  & \textcolor{gray}{\textit{10.38}} & \cellcolor{azureblue}\textcolor{gray}{\textit{9.93}}  & \textcolor{gray}{\textit{1131}} & \textcolor{gray}{\textit{49.0}} \\
Qwen2.5-7B-DPO           & 3.92          & 24.22          & \cellcolor{azureblue}12.04          & 25.05          & 15.13          & 15.96          & 18.39          & \cellcolor{azureblue}17.01          & 1420          & 94.0          \\
Qwen2.5-7B-cDPO          & 3.05          & 25.80          & \cellcolor{azureblue}12.15          & 25.20          & 15.25          & 14.55          & 17.65          & \cellcolor{azureblue}16.24          & 1303          & 96.0          \\
Qwen2.5-7B-IPO           & 2.94          & 27.14          & \cellcolor{azureblue}12.62          & \textbf{26.40} & \textbf{17.83} & \textbf{18.42} & 19.22          & \cellcolor{azureblue}\textbf{18.70} & 1521          & \textbf{98.7}          \\
Qwen2.5-7B-SimPO         & 1.39          & \textbf{30.67} & \cellcolor{azureblue}13.10          & 25.42          & 15.32          & 15.32          & 17.71          & \cellcolor{azureblue}16.52          & 1561          & 97.8          \\
Qwen2.5-7B-GM-DPO (Ours) & \textbf{6.14} & 25.95          & \cellcolor{azureblue}\textbf{14.06} & 25.82          & 17.05          & 17.31          & \textbf{19.72} & \cellcolor{azureblue}18.46          & 1506          & 98.2 \\
\midrule
\textcolor{gray}{\textit{Qwen3-8B-Base}}     & \textcolor{gray}{\textit{5.85}}  & \textcolor{gray}{\textit{40.25}} & \cellcolor{azureblue}\textcolor{gray}{\textit{19.61}} & \textcolor{gray}{\textit{25.67}} & \textcolor{gray}{\textit{13.09}} & \textcolor{gray}{\textit{12.75}} & \textcolor{gray}{\textit{15.99}} & \cellcolor{azureblue}\textcolor{gray}{\textit{14.44}} & \textcolor{gray}{\textit{1391}} & \textcolor{gray}{\textit{37.2}} \\
Qwen3-8B-DPO             & 7.16          & 38.40          & \cellcolor{azureblue}19.66          & 28.86          & 17.69          & 19.94          & 20.52          & \cellcolor{azureblue}19.78          & 1572          & 55.6          \\
Qwen3-8B-cDPO            & 6.44          & 33.76          & \cellcolor{azureblue}17.37          & 27.37          & 16.73          & 17.32          & 20.25          & \cellcolor{azureblue}18.67          & 1514          & 63.7          \\
Qwen3-8B-IPO             & 7.05          & 39.33          & \cellcolor{azureblue}19.96          & 29.08          & 18.35          & 19.19          & 21.74          & \cellcolor{azureblue}20.30          & 1563          & 86.3          \\
Qwen3-8B-SimPO           & 6.64          & 39.58          & \cellcolor{azureblue}19.82          & 28.60          & 18.84          & 19.77          & 21.35          & \cellcolor{azureblue}20.37          & 1615          & 77.7          \\
Qwen3-8B-GM-DPO (Ours)   & \textbf{9.30} & \textbf{43.69} & \cellcolor{azureblue}\textbf{23.06} & \textbf{30.52} & \textbf{21.81} & \textbf{20.58} & \textbf{23.60} & \cellcolor{azureblue}\textbf{22.34} & 1610          & \textbf{89.8} \\
\midrule
\textcolor{gray}{\textit{LLaVA-2-8B-Base}}   & \textcolor{gray}{\textit{4.24}}  & \textcolor{gray}{\textit{48.70}} & \cellcolor{azureblue}\textcolor{gray}{\textit{22.02}} & \textcolor{gray}{\textit{23.21}} & \textcolor{gray}{\textit{4.94}}  & \textcolor{gray}{\textit{5.79}}  & \textcolor{gray}{\textit{6.38}}  & \cellcolor{azureblue}\textcolor{gray}{\textit{5.92}}  & \textcolor{gray}{\textit{605}}  & \textcolor{gray}{\textit{2.2}}  \\
LLaVA-2-8B-DPO           & 10.66         & 40.73          & \cellcolor{azureblue}22.69          & 29.15          & 21.17          & 22.53          & 23.33          & \cellcolor{azureblue}22.66          & 1320          & 25.3          \\
LLaVA-2-8B-cDPO          & 6.35          & 45.46          & \cellcolor{azureblue}21.99          & 25.54          & 13.44          & 13.58          & 14.90          & \cellcolor{azureblue}14.21          & 1298          & 36.9          \\
LLaVA-2-8B-IPO           & 10.24         & 41.40          & \cellcolor{azureblue}22.70          & 29.22          & 22.89          & 23.00          & 24.01          & \cellcolor{azureblue}23.48          & 1413          & 39.5          \\
LLaVA-2-8B-SimPO         & 6.93          & \textbf{47.42} & \cellcolor{azureblue}23.13          & 26.56          & 16.90          & 18.19          & 15.71          & \cellcolor{azureblue}16.69          & 1400          & 32.6          \\
LLaVA-2-8B-GM-DPO (Ours) & \textbf{13.89}& 42.66          & \cellcolor{azureblue}\textbf{25.40} & \textbf{31.77} & \textbf{25.87} & \textbf{25.49} & \textbf{27.04} & \cellcolor{azureblue}\textbf{26.34} & 1405          & \textbf{44.7} \\
\bottomrule
\end{tabular}
}
\end{center}
\vspace{-1.6em}
\end{table*}

\vspace{-1em}
\subsection{Main Alignment Results}
\label{sec:main_alignment}
\vspace{-0.1em}

We next examine whether targeted preference optimization can
address these limitations while retaining dense caption
output. 
As shown in Table~\ref{tab:main_alignment_results},
we compare preference objectives across three backbones
with the Qwen3-8B SFT baseline serving as a reference for caption imitation alone.

\textbf{Fine-Grained Action Grounding.}
Standard DPO treats minor omissions identically to fatal chirality inversions, resulting in poor performance under evaluation systems that penalize hallucinations.
GM-DPO incorporates this information through severity-dependent margins and weights, achieving the highest Substantive scores among
the evaluated preference methods on all three backbones. 
Relative to DPO, scores increase from 3.92 to 6.14
on Qwen2.5-7B, 7.16 to 9.30 on Qwen3-8B, and
10.66 to 13.89 on LLaVA-2-8B.
Non-substantive scores also improve over DPO on every
backbone. Although other objectives obtain higher
non-substantive scores on Qwen2.5-7B and LLaVA-2-8B,
GM-DPO consistently leads on the substantive actions
targeted by our preference construction.

\textbf{Robustness under Penalty-Aware Evaluation.}
To examine whether these improvements persist when incorrect assertions incur penalties, we compare the aggregate GPA scores. 
GM-DPO ranks first among the evaluated alignment methods on every backbone, reaching 14.06, 23.06, and 25.40, respectively, with gains of 2.02, 3.40, and 2.71 points over DPO. 
These gains contain two measurable contributions: $\mathrm{GPA}_{\mathrm{clip}}$ increases by 0.77, 1.02, and 2.62 points, while the aggregate penalty deduction, $\mathrm{GPA}_{\mathrm{clip}}-\mathrm{GPA}$, decreases from 13.01 to 11.76, 9.20 to 7.46, and 6.46 to 6.37. The improvement therefore combines higher positive credit with smaller penalties, with their relative contributions varying across backbones. In particular, the LLaVA-2-8B gain primarily reflects increased positive credit rather than a large reduction in penalties.

\textbf{Consecutive-Action Coverage across Shots.}
Sequence queries jointly evaluate consecutive substantive
actions in their corresponding shot contexts, connecting
individual action fidelity with coverage across shots.
Compared to DPO, GM-DPO consistently improves $\mathrm{TSA}_{\mathrm{w}}$ by 1.45, 2.56, and 3.68 points, respectively.
It achieves the highest $\mathrm{Seq}_{\mathrm{chain}}$
on all three backbones and leads all sequence metrics
on Qwen3-8B and LLaVA-2-8B.
On Qwen2.5-7B, IPO retains a slightly higher
$\mathrm{TSA}_{\mathrm{w}}$ of 18.70.
These results support improved joint coverage of
consecutive actions; because the queries share action
requirements, they do not isolate temporal reasoning
from action accuracy.

\textbf{Maintaining Caption Length and Shot Structure.}
A common failure mode in preference optimization is length or format collapse, where models generate truncated answers to avoid penalties. As shown in Table~\ref{tab:main_alignment_results}, all preference-tuned variants preserve descriptive length while improving SMR over base models. Notably, on Qwen3-8B, GM-DPO matches full-SFT in both length (1610 vs. 1634) and structure compliance (89.8\% vs. 89.7\% SMR), demonstrating that our auxiliary NLL loss prevents linguistic collapse without standalone SFT. Nevertheless, final SMR gains remain largely bounded by each backbone's inherent instruction-following capacity.

\subsection{Diagnostic Analysis and Evaluator Fidelity}
\label{sec:diagnostic_analysis}
\begin{figure*}[t]
\centering
\includegraphics[width=0.993\linewidth] {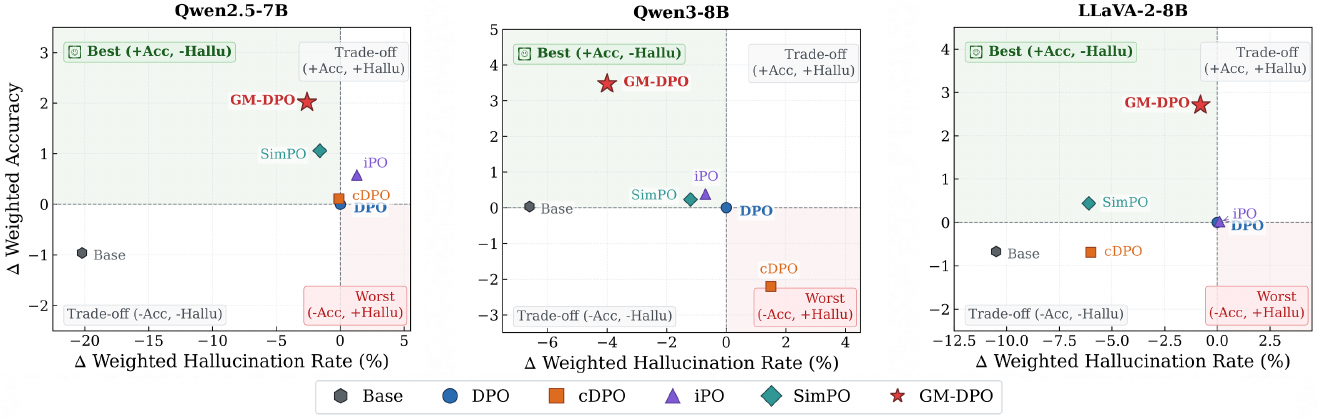}
\vspace{-0.8em}
\caption{The accuracy-hallucination trade-off across three VLM architectures on FlexBench, normalized relative to standard DPO as origin (0, 0).}
\label{fig:pareto_frontier}
\vspace{-0.2em}
\end{figure*}

\begin{figure*}[t]
\centering
\includegraphics[width=1\linewidth]{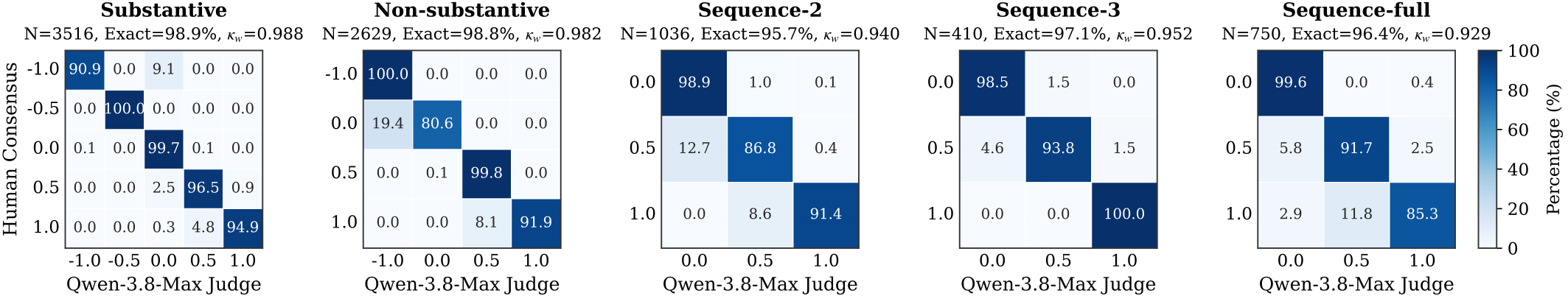}
\vspace{-1.8em}
\caption{Row-normalized agreement matrices between human consensus and LLM.}
\label{fig:evaluator_agreement_matrices}
\vspace{-1.8em}
\end{figure*}

\textbf{Physical Fidelity and Hallucination.}
As shown in Figure~\ref{fig:pareto_frontier}, we compare changes in GPA and WHR relative to DPO to examine how physical alignment improvements relate to hallucination penalties. GM-DPO occupies the upper-left region on all three backbones, achieving the largest GPA increase among the compared alignment methods while also reducing WHR. Several baselines also reach this region, so the advantage is not exclusive quadrant membership. For example, SimPO achieves a larger WHR reduction on LLaVA-2-8B, but its GPA gain is smaller than GM-DPO's (0.44 versus 2.71 points). On Qwen3-8B, GM-DPO reduces WHR from 18.8\% to 14.8\%, a reduction of 4.0 percentage points, while increasing GPA by 3.40 points. These results show that improved penalty-aware fidelity can accompany reduced hallucination rates, with the balance varying across objectives and backbones. Representative caption comparisons are provided in Appendix~\ref{sec:appendix_qualitative}.

\textbf{Evaluator Reliability and Human Agreement.}
To validate the reliability of FlexBench's automated evaluation protocol, we benchmark the LLM judge (\texttt{Qwen-3.8-Max}, $T=0.0$) against consensus annotations from 20 trained experts across 8,341 query instances generated by 28 diverse models. As shown in Figure~\ref{fig:evaluator_agreement_matrices}, exact agreement spans from 95.7\% to 98.9\% across all five suites, accompanied by outstanding quadratic weighted $\kappa_w$ values ranging from 0.929 to 0.988. By demonstrating robust alignment across a broad spectrum of model outputs, these results confirm the high fidelity, objectivity, and practical viability of FlexBench's automated protocol as a reliable proxy for human judgment in fine-grained video evaluation.

\subsection{Ablation on GM-DPO Loss Formulation}
\label{sec:ablation_study}

\begin{wraptable}{r}{0.36\linewidth}
\vspace{-2.2em}
\centering
\caption{Ablation across backbones.}
\label{tab:ablation_compact}
\vspace{0.15em}
\scriptsize
\setlength{\tabcolsep}{3pt}
\renewcommand{\arraystretch}{1.10}
\begin{tabular}{l|cc|cc|c}
\toprule
\textbf{Model} & $\omega$ & $m$ & \textbf{GPA}
& $\mathbf{GPA}_{\mathrm{clip}}$
& $\mathbf{TSA}_{\mathrm{w}}$ \\
\midrule
\multirow{4}{*}{Qwen2.5}
& \xmark & \xmark & 12.04 & 25.05 & 17.01 \\
& \cmark & \xmark & 12.29 & 25.00 & 16.52 \\
& \xmark & \cmark & 13.20 & 25.44 & 17.73 \\
& \cmark & \cmark & \textbf{14.06} & \textbf{25.82} & \textbf{18.46} \\
\midrule
\multirow{4}{*}{Qwen3}
& \xmark & \xmark & 19.66 & 28.86 & 19.78 \\
& \cmark & \xmark & 19.93 & 29.42 & 19.88 \\
& \xmark & \cmark & 20.07 & 29.82 & 20.03 \\
& \cmark & \cmark & \textbf{23.06} & \textbf{30.52} & \textbf{22.34} \\
\midrule
\multirow{4}{*}{LLaVA-2}
& \xmark & \xmark & 22.69 & 29.15 & 22.66 \\
& \cmark & \xmark & 23.77 & 28.97 & 23.27 \\
& \xmark & \cmark & 24.55 & 30.63 & 25.27 \\
& \cmark & \cmark & \textbf{25.40} & \textbf{31.77} & \textbf{26.34} \\
\bottomrule
\end{tabular}
\vspace{-0.9em}
\end{wraptable}
We ablate the severity-dependent weight ($\omega$)
and margin ($m$), recovering DPO when both are disabled
while retaining the auxiliary NLL loss.
As shown in Table~\ref{tab:ablation_compact}, the margin
alone improves GPA by 1.16, 0.41, and 1.86 points
on Qwen2.5, Qwen3, and LLaVA-2, respectively,
outperforming weighting alone on every backbone.
Weighting alone also improves GPA, but its benefits
do not consistently extend to credited content
or sequence coverage.
Combining both components yields the highest GPA,
$\mathrm{GPA}_{\mathrm{clip}}$, and
$\mathrm{TSA}_{\mathrm{w}}$ in every ablation group,
supporting their complementary roles in improving
action fidelity and consecutive-action coverage.
This benefit is clearest on Qwen3, where joint
optimization reaches 23.06 GPA, compared with
19.93 for weighting alone and 20.07 for the margin alone.
Detailed results appear in
Appendix~\ref{sec:appendix_full_ablation}.

\subsection{Generalization to Downstream Multimodal Benchmarks}
\label{sec:generalization}

\begin{wraptable}{r}{0.37\linewidth}
\vspace{-2.2em}
\centering
\caption{Cross-benchmark generalization across backbones.}
\label{tab:generalization_compact}
\vspace{0.2em}
\scriptsize
\setlength{\tabcolsep}{1.8pt}
\renewcommand{\arraystretch}{1.08}
\begin{tabular}{l|c|cc|c}
\toprule
\multirow{2}{*}{\textbf{Method Variant}}
& \textbf{V-MME}
& \multicolumn{2}{c|}{\textbf{POPE}}
& \textbf{V-GPT} \\
& \textbf{Acc} & \textbf{Acc} & \textbf{F1} & \textbf{Score} \\
\midrule
Q2.5-DPO
& 60.06 & 87.43 & 86.06 & \underline{2.50} \\
Q2.5-cDPO
& 59.58 & \underline{87.46} & 86.06 & \underline{2.50} \\
Q2.5-IPO
& \underline{60.45} & 87.44 & \underline{86.09} & 2.46 \\
Q2.5-SimPO
& 60.26 & 87.41 & 86.02 & 2.48 \\
\textbf{Q2.5-GM-DPO}
& \textbf{60.62} & \textbf{87.56} & \textbf{86.15} & \textbf{2.51} \\
\midrule
Q3-DPO
& 62.35 & \underline{89.37} & \underline{88.98} & 2.70 \\
Q3-cDPO
& \textbf{62.73} & \underline{89.37} & \underline{88.98}
& \underline{2.73} \\
Q3-IPO
& 62.07 & \underline{89.37} & 88.96 & 2.66 \\
Q3-SimPO
& 62.24 & 89.35 & 88.94 & 2.64 \\
\textbf{Q3-GM-DPO}
& \underline{62.72} & \textbf{89.40} & \textbf{89.02}
& \textbf{2.75} \\
\midrule
L2-DPO
& 64.67 & 88.78 & 88.31 & 2.51 \\
L2-cDPO
& 65.14 & \underline{88.88} & \underline{88.34}
& \underline{2.56} \\
L2-IPO
& \underline{65.19} & 88.77 & 88.11 & \underline{2.56} \\
L2-SimPO
& 64.78 & 88.49 & 88.04 & 2.55 \\
\textbf{L2-GM-DPO}
& \textbf{65.39} & \textbf{88.97} & \textbf{88.41}
& \textbf{2.57} \\
\bottomrule
\end{tabular}
\vspace{-0.9em}
\end{wraptable}

To examine whether fine-grained motion alignment retains broader multimodal capabilities, we evaluate the aligned checkpoints on Video-MME~\citep{fu2024videomme}, POPE~\citep{li2023evaluating}, and Video-ChatGPT~\citep{maaz2024videochatgpt}. As shown in Table~\ref{tab:generalization_compact}, GM-DPO achieves the highest POPE accuracy and F1, together with the highest Video-ChatGPT overall score, among the compared alignment methods on all three backbones. It also leads Video-MME on Qwen2.5 and LLaVA-2; on Qwen3, its accuracy of 62.72 is slightly below cDPO's 62.73.
Crucially, cross-benchmark comparison reveals an overarching diagnostic limitation of prevailing suites: performance deltas across disparate alignment objectives remain heavily compressed (fluctuating within $\pm 0.1$ on Video-ChatGPT), as coarse event recognition and multiple-choice probes lack sensitivity to granular physical dynamics. In stark contrast, FlexBench exposes decisive gaps across these identical models (spanning up to 3.40 GPA points and a 21.3\% WHR delta in Section~\ref{sec:diagnostic_analysis}), confirming metric saturation on standard benchmarks and highlighting FlexBench as an indispensable diagnostic testbed. Comprehensive breakdowns are provided in Appendix~\ref{sec:appendix_generalization}.

\section{Conclusion}
\label{sec:conclusion}

We presented a framework combining fine-grained
evaluation and severity-aware preference learning
to improve limb-motion fidelity in dense video captions.
FlexBench systematically evaluates each identifiable
person's actions and limb states across shots,
while GPA distinguishes credited content from
incorrect claims through graded scores and penalties.
GM-DPO brings this rubric into preference margins
and loss weights, consistently improving
substantive-action fidelity and consecutive-action
coverage over DPO across three backbones and
training seeds.
These gains accompany sustained long-form output
and competitive performance on broader multimodal
benchmarks, supporting targeted motion alignment
within general-purpose captioning models.
However, monocular ambiguity remains a challenge,
and these results do not establish the accuracy
of other caption content, such as camera angles,
camera motion, or sound.
Future work will evaluate these dimensions and
examine whether more accurate motion captions
provide better supervision for downstream learning.
Further limitations and research directions are
discussed in Appendix~\ref{sec:limitations}.

\bibliography{iclr2027_conference}
\bibliographystyle{iclr2027_conference}

\appendix

\section*{Appendix Overview}

The appendix is organized into four sections.
Section~\ref{sec:appendix_data_curation} details
FlexBench video selection, annotation, and quality
assurance, together with perturbation protocols
and human assessment of caption naturalness.
Section~\ref{sec:appendix_training_details} documents
the training setup, random-seed settings, baseline
objectives, and method-specific hyperparameters.
Section~\ref{sec:appendix_results} begins with
results across three random seeds
(Section~\ref{sec:appendix_seed_robustness}),
followed by complete ablation results (Section~\ref{sec:appendix_full_ablation}), evaluations
on Video-MME, POPE, and Video-ChatGPT (Section~\ref{sec:appendix_generalization}), and
qualitative caption comparisons (Section~\ref{sec:appendix_qualitative}).
Section~\ref{sec:limitations} discusses limitations
and future directions, including extensions of
the grading rubric, ambiguity in monocular video,
and broader evaluation of dense caption content.

\section{FlexBench Curation and Annotation Details}
\label{sec:appendix_data_curation}

\subsection{Video Filtering and Sequence Curation}
Raw footage harvested from the AVA repository~\citep{gu2018ava} and YouTube Creative Commons streams was re-segmented and concatenated to build an initial pool of over 56k candidate sequences. We applied automated heuristic screening to filter out sequences containing shots shorter than 1.5\,s, instances with minimal physical motion, scenery-dominated sequences, and sensitive or unsafe footage, reducing the pool to 12k sequences. Subsequently, domain experts conducted stratified manual selection prioritizing scenario diversity, actor density, and rich limb-level physical interactions. This curation yielded the final FlexBench benchmark comprising 660 verified video sequences spanning 3,105 distinct continuous shots.

\subsection{Human-in-the-Loop Annotation and Multi-Round Verification}
\label{sec:human_annotation}
Following automated VLM pre-captioning, an expert panel of 18 trained annotators holding bachelor’s degrees refined candidate annotations under a standardized rubric covering cross-shot subject tracking, visual appearance, interacted affordances, limb chirality (explicitly tagging left, right, or bimanual execution), and chronological connectives. Annotations underwent three rounds of blind cross-verification. Raw inter-annotator agreement was 94.2\%, with Fleiss' $\kappa=0.89$. For the remaining edge cases—primarily involving extreme perspective foreshortening or partial self-occlusions—chirality and action states were finalized via panel consensus arbitration based on a two-thirds majority voting protocol under multi-frame zoom examination. Finally, atomic positive QA probes were derived from the verified captions, with every probe independently audited across two rounds of human inspection to ensure precise physical factuality.

\subsection{Linguistic Naturalness and Non-Spuriousness of Preference Pairs}
\label{sec:perturbation_validation}

\textbf{Locally Constrained Perturbation Protocols.}
To eliminate textual distribution discrepancies between preferred and dispreferred pairs, all perturbations are strictly confined to local limb kinematics, keeping the surrounding sentence structures, character appearances, and stylistic register completely frozen. As introduced in Section~\ref{sec:pair_generation}, directional laterality is modified via deterministic token substitution (e.g., swapping \textit{``left''} and \textit{``right''} hands or stripping laterality qualifiers), while temporal omissions, subtle subaction variations, and phantom actions are synthesized using Gemini-3.5-Flash under rigid schema constraints. Because ground-truth captions ($y_w$) are originally initialized by frontier VLMs before human refinement, both $y_w$ and $y_l$ share identical underlying linguistic foundations and syntactic distributions.

\begin{figure}[t]
    \centering
    \includegraphics[width=0.85\linewidth]{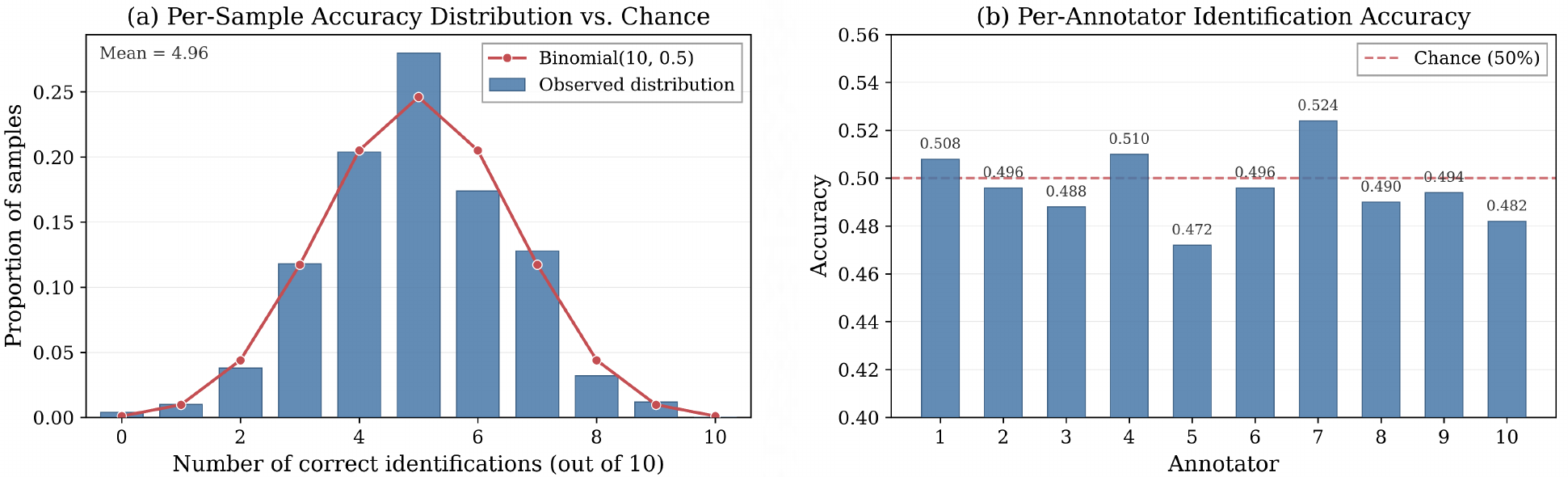} 
    \caption{Double-blind human identification study on preference pair naturalness. (a) The empirical distribution of correct identifications per sample closely matches the theoretical $\text{Binomial}(10, 0.5)$ chance baseline. (b) Individual identification accuracies across all 10 human evaluators fluctuate tightly around the 50\% chance threshold (overall accuracy = 49.60\%, binomial $p = 0.5813 > 0.05$).}
    \label{fig:human_blind_test}
\end{figure}

\textbf{Double-Blind Human Indistinguishability Evaluation.}
To verify that models optimize genuine physical grounding rather than discriminating between human-edited and LLM-synthesized writing styles, we conducted a double-blind identification experiment. Ten independent annotators with no prior exposure to the dataset evaluated 500 randomly sampled blind pairs (5,000 total judgments) to identify human-refined versus LLM-modified descriptions. As shown in Figure~\ref{fig:human_blind_test}, human evaluators achieved an overall accuracy of 49.60\% (virtually identical to random chance; binomial test $p = 0.5813 > 0.05$), with both sample-level and individual identification distributions conforming closely to the theoretical $\text{Binomial}(10, 0.5)$ null distribution. This confirms that our targeted kinematic modifications introduce no stylistic artifacts or distributional drift, ensuring that GM-DPO is driven strictly by visual-physical motion perception.

\section{Experimental Setup and Baseline Objectives}
\label{sec:appendix_training_details}

All preference alignment experiments are conducted using the \texttt{ms-swift} framework. We perform full-parameter instruction tuning on the language backbones while freezing both the vision encoders and cross-modal projector modules. Training is executed in \texttt{bfloat16} precision using the AdamW optimizer with a learning rate of $5\times 10^{-7}$. The global batch size is 128. Across all alignment objectives, an auxiliary supervised next-token prediction loss ($+1.0 \times \mathcal{L}_{\text{NLL}}$) is incorporated to preserve generation fluency. Given preference tuples $\mathcal{D} = (x, y_w, y_l)$, Table~\ref{tab:appendix_loss_objectives} summarizes the mathematical formulations and key hyperparameter configurations across the evaluated offline alignment methods. For both SFT and preference optimization, we conduct
three training runs with
$(\texttt{seed},\texttt{data\_seed})$
set to $(42,42)$, $(106,106)$, and $(298,298)$.
The global seed controls stochastic training operations,
while \texttt{data\_seed} controls data shuffling.
Both seeds vary together across runs, and reported
results for trained models are arithmetic means over
the three runs.

\begin{table}[t]
\centering
\small
\caption{Mathematical formulations and key hyperparameter configurations of benchmarked offline preference optimization objectives. Let $\Delta r_\theta = \beta \log \frac{\pi_\theta(y_w \mid x)}{\pi_{\text{ref}}(y_w \mid x)} - \beta \log \frac{\pi_\theta(y_l \mid x)}{\pi_{\text{ref}}(y_l \mid x)}$ denote the implicit reward margin. All methods incorporate an auxiliary $+1.0 \times \mathcal{L}_{\text{NLL}}$ loss.}
\label{tab:appendix_loss_objectives}
\renewcommand{\arraystretch}{1.35}
\setlength{\tabcolsep}{3.8pt}
\resizebox{\linewidth}{!}{
\setlength{\tabcolsep}{14.6pt}
\begin{tabular}{lll}
\toprule
\textbf{Method}
& \textbf{Preference Loss} $\ell_M$
& \textbf{Hyperparameters} \\
\midrule
\shortstack[l]{DPO ~\citep{rafailov2023direct}}
& $-\log\sigma(\Delta r_\theta)$
& $\beta=0.1$ \\

\shortstack[l]{cDPO ~\citep{mitchell2023cdpo}}
& $-(1-\varepsilon)\log\sigma(\Delta r_\theta)
   -\varepsilon\log\sigma(-\Delta r_\theta)$
& $\beta=0.1,\ \varepsilon=0.1$ \\

\shortstack[l]{IPO ~\citep{azar2023general}}
& $\left(\bar{h}_\theta-\frac{1}{2\beta}\right)^2$
& $\beta=0.5$ \\

\shortstack[l]{SimPO ~\citep{meng2024simpo}}
& $-\log\sigma\!\left(
   \beta[q_\theta(y_w\mid x)-q_\theta(y_l\mid x)]
   -\gamma_{\mathrm{S}}\right)$
& $\beta=2.0,\ \gamma_{\mathrm{S}}=1.0$ \\
\midrule
\textbf{GM-DPO (Ours)}
& $-(1+\alpha\Delta S)
   \log\sigma(\Delta r_\theta-\gamma\Delta S)$
& $\beta=0.1,\ \gamma=0.2,\ \alpha=0.1$ \\
\bottomrule
\end{tabular}
}
\end{table}

\section{Comprehensive Experimental Results and Visualizations}
\label{sec:appendix_results}

\subsection{Robustness across Random Seeds}
\label{sec:appendix_seed_robustness}

We repeat each preference method on all three backbones
and SFT on Qwen3-8B using the three seed configurations
in Appendix~\ref{sec:appendix_training_details}, jointly
varying the training and data-order seeds while fixing
all other settings.
Table~\ref{tab:seed_robustness} reports means and sample
standard deviations ($n=3$, denominator $n-1$), with
the highest mean per backbone in bold; the means match
Table~\ref{tab:main_alignment_results}.
GM-DPO leads the evaluated preference methods in
substantive-action accuracy, GPA, and
$\mathrm{Seq}_{\mathrm{chain}}$ on every backbone
in all three runs.
It also consistently leads in
$\mathrm{GPA}_{\mathrm{clip}}$ and
$\mathrm{TSA}_{\mathrm{w}}$ on Qwen3-8B and LLaVA-2-8B,
and exceeds SFT on Qwen3-8B in mean substantive-action
accuracy (9.30 versus 6.92) and GPA (23.06 versus 21.08).
On Qwen2.5-7B, IPO retains higher mean
$\mathrm{GPA}_{\mathrm{clip}}$ and
$\mathrm{TSA}_{\mathrm{w}}$, with the TSA ranking
between IPO and GM-DPO varying across runs.

\begin{table*}[t]
\centering
\caption{FlexBench results across three training seeds, reported as mean $\pm$ sample standard deviation. The highest mean within each backbone is highlighted in \textbf{bold}.}
\label{tab:seed_robustness}
\vspace{0.3em}
\resizebox{\linewidth}{!}{%
\setlength{\tabcolsep}{3pt}
\renewcommand{\arraystretch}{1.12}
\begin{tabular}{l|cccc|cccc}
\toprule
\multicolumn{1}{c|}{\textbf{Method}}
& \multicolumn{4}{c|}{\textbf{Motion Accuracy}}
& \multicolumn{4}{c}{\textbf{Motion Sequence}} \\
& \textbf{Subs.}
& \textbf{Non-subs.}
& \cellcolor{azureblue}\textbf{GPA}
& $\mathbf{GPA}_{\mathrm{clip}}$
& $\mathbf{Seq}_{2}$
& $\mathbf{Seq}_{3}$
& $\mathbf{Seq}_{\mathrm{chain}}$
& \cellcolor{azureblue}$\mathbf{TSA}_{\mathrm{w}}$ \\
\midrule

\multicolumn{9}{l}{\textit{\textcolor{gray}{Qwen2.5-7B}}} \\
DPO
& $3.92 \pm 0.12$
& $24.22 \pm 0.29$
& \cellcolor{azureblue}$12.04 \pm 0.11$
& $25.05 \pm 0.29$
& $15.13 \pm 0.34$
& $15.96 \pm 0.13$
& $18.39 \pm 0.21$
& \cellcolor{azureblue}$17.01 \pm 0.18$ \\
cDPO
& $3.05 \pm 0.33$
& $25.80 \pm 0.13$
& \cellcolor{azureblue}$12.15 \pm 0.25$
& $25.20 \pm 0.29$
& $15.25 \pm 0.45$
& $14.55 \pm 0.33$
& $17.65 \pm 0.25$
& \cellcolor{azureblue}$16.24 \pm 0.13$ \\
IPO
& $2.94 \pm 0.33$
& $27.14 \pm 0.14$
& \cellcolor{azureblue}$12.62 \pm 0.18$
& $\mathbf{26.40} \pm 0.27$
& $\mathbf{17.83} \pm 0.39$
& $\mathbf{18.42} \pm 0.33$
& $19.22 \pm 0.20$
& \cellcolor{azureblue}$\mathbf{18.70} \pm 0.25$ \\
SimPO
& $1.39 \pm 0.39$
& $\mathbf{30.67} \pm 0.13$
& \cellcolor{azureblue}$13.10 \pm 0.20$
& $25.42 \pm 0.18$
& $15.32 \pm 0.25$
& $15.32 \pm 0.22$
& $17.71 \pm 0.42$
& \cellcolor{azureblue}$16.52 \pm 0.25$ \\
\textbf{GM-DPO}
& $\mathbf{6.14} \pm 0.27$
& $25.95 \pm 0.31$
& \cellcolor{azureblue}$\mathbf{14.06} \pm 0.05$
& $25.82 \pm 0.22$
& $17.05 \pm 0.31$
& $17.31 \pm 0.17$
& $\mathbf{19.72} \pm 0.31$
& \cellcolor{azureblue}$18.46 \pm 0.25$ \\
\midrule

\multicolumn{9}{l}{\textit{\textcolor{gray}{Qwen3-8B}}} \\
SFT
& $6.92 \pm 0.29$
& $42.31 \pm 0.14$
& \cellcolor{azureblue}$21.08 \pm 0.14$
& $29.88 \pm 0.12$
& $13.30 \pm 0.14$
& $15.54 \pm 0.20$
& $15.81 \pm 0.35$
& \cellcolor{azureblue}$15.23 \pm 0.23$ \\
DPO
& $7.16 \pm 0.33$
& $38.40 \pm 0.46$
& \cellcolor{azureblue}$19.66 \pm 0.31$
& $28.86 \pm 0.24$
& $17.69 \pm 0.36$
& $19.94 \pm 0.28$
& $20.52 \pm 0.24$
& \cellcolor{azureblue}$19.78 \pm 0.14$ \\
cDPO
& $6.44 \pm 0.36$
& $33.76 \pm 0.14$
& \cellcolor{azureblue}$17.37 \pm 0.16$
& $27.37 \pm 0.39$
& $16.73 \pm 0.17$
& $17.32 \pm 0.14$
& $20.25 \pm 0.23$
& \cellcolor{azureblue}$18.67 \pm 0.08$ \\
IPO
& $7.05 \pm 0.15$
& $39.33 \pm 0.42$
& \cellcolor{azureblue}$19.96 \pm 0.24$
& $29.08 \pm 0.18$
& $18.35 \pm 0.25$
& $19.19 \pm 0.34$
& $21.74 \pm 0.25$
& \cellcolor{azureblue}$20.30 \pm 0.12$ \\
SimPO
& $6.64 \pm 0.29$
& $39.58 \pm 0.41$
& \cellcolor{azureblue}$19.82 \pm 0.30$
& $28.60 \pm 0.18$
& $18.84 \pm 0.29$
& $19.77 \pm 0.18$
& $21.35 \pm 0.21$
& \cellcolor{azureblue}$20.37 \pm 0.21$ \\
\textbf{GM-DPO}
& $\mathbf{9.30} \pm 0.30$
& $\mathbf{43.69} \pm 0.28$
& \cellcolor{azureblue}$\mathbf{23.06} \pm 0.16$
& $\mathbf{30.52} \pm 0.21$
& $\mathbf{21.81} \pm 0.46$
& $\mathbf{20.58} \pm 0.38$
& $\mathbf{23.60} \pm 0.37$
& \cellcolor{azureblue}$\mathbf{22.34} \pm 0.17$ \\
\midrule

\multicolumn{9}{l}{\textit{\textcolor{gray}{LLaVA-2-8B}}} \\
DPO
& $10.66 \pm 0.31$
& $40.73 \pm 0.20$
& \cellcolor{azureblue}$22.69 \pm 0.19$
& $29.15 \pm 0.28$
& $21.17 \pm 0.26$
& $22.53 \pm 0.24$
& $23.33 \pm 0.35$
& \cellcolor{azureblue}$22.66 \pm 0.23$ \\
cDPO
& $6.35 \pm 0.36$
& $45.46 \pm 0.39$
& \cellcolor{azureblue}$21.99 \pm 0.34$
& $25.54 \pm 0.15$
& $13.44 \pm 0.29$
& $13.58 \pm 0.31$
& $14.90 \pm 0.40$
& \cellcolor{azureblue}$14.21 \pm 0.30$ \\
IPO
& $10.24 \pm 0.34$
& $41.40 \pm 0.28$
& \cellcolor{azureblue}$22.70 \pm 0.17$
& $29.22 \pm 0.28$
& $22.89 \pm 0.40$
& $23.00 \pm 0.38$
& $24.01 \pm 0.28$
& \cellcolor{azureblue}$23.48 \pm 0.23$ \\
SimPO
& $6.93 \pm 0.27$
& $\mathbf{47.42} \pm 0.09$
& \cellcolor{azureblue}$23.13 \pm 0.15$
& $26.56 \pm 0.16$
& $16.90 \pm 0.17$
& $18.19 \pm 0.33$
& $15.71 \pm 0.32$
& \cellcolor{azureblue}$16.69 \pm 0.23$ \\
\textbf{GM-DPO}
& $\mathbf{13.89} \pm 0.24$
& $42.66 \pm 0.32$
& \cellcolor{azureblue}$\mathbf{25.40} \pm 0.15$
& $\mathbf{31.77} \pm 0.07$
& $\mathbf{25.87} \pm 0.29$
& $\mathbf{25.49} \pm 0.44$
& $\mathbf{27.04} \pm 0.27$
& \cellcolor{azureblue}$\mathbf{26.34} \pm 0.22$ \\
\bottomrule
\end{tabular}%
}
\vspace{-0.6em}
\end{table*}


\begin{wraptable}{r}{0.5\linewidth}
\vspace{-1.6em}
\centering
\caption{\small Paired GPA gains over DPO.}
\label{tab:paired_seed_gains}
\vspace{0.2em}
\scriptsize
\setlength{\tabcolsep}{2.5pt}
\renewcommand{\arraystretch}{1.10}
\resizebox{\linewidth}{!}{%
\begin{tabular}{l|ccc|cc}
\toprule
\textbf{Backbone}
& \multicolumn{3}{c|}{\textbf{GPA Gain}}
& \textbf{Mean}
& \textbf{95\% CI} \\
& \textbf{Run 1}
& \textbf{Run 2}
& \textbf{Run 3}
& & \\
\midrule
{Qwen2.5-7B}
& $+2.12$ & $+1.83$ & $+2.11$
& $+2.02$ & $[1.61,\,2.43]$ \\
{Qwen3-8B}
& $+3.41$ & $+3.80$ & $+2.99$
& $+3.40$ & $[2.39,\,4.41]$ \\
{LLaVA-2-8B}
& $+2.57$ & $+2.67$ & $+2.90$
& $+2.71$ & $[2.29,\,3.13]$ \\
\bottomrule
\end{tabular}%
}
\vspace{-0.6em}
\end{wraptable}

\noindent\textbf{Paired Improvements over DPO.}
We pair runs with identical training and data-order
seed settings and compute the within-pair GPA
difference $d_i$ between GM-DPO and DPO.
Table~\ref{tab:paired_seed_gains} reports these gains
and pointwise 95\% confidence intervals (CIs),
calculated as $\bar{d}\pm t_{0.975,2}s_d/\sqrt{3}$,
where $s_d$ is the sample standard deviation of
the paired differences.
The intervals use two degrees of freedom and no
multiple-comparison adjustment.
All paired gains are positive, and all intervals
lie above zero.
These intervals summarize cross-run uncertainty
on the fixed evaluation set under independent,
approximately normal paired differences.
With only three runs, this assumption cannot be
reliably checked; the intervals therefore provide
supplementary evidence alongside the consistently
positive observed gains.

\subsection{Full Fine-Grained Kinematic and Sequential Ablation}
\label{sec:appendix_full_ablation}

Table~\ref{tab:appendix_full_ablation} extends the
main-paper ablation to all action and sequence metrics.
We separately evaluate the severity-dependent margin
and loss weight, retaining the same auxiliary NLL
term across variants.
Disabling both components recovers the DPO baseline.

\textbf{Effect of the Severity-Dependent Margin.}
The margin-only variant improves substantive-action
scores from 3.92 to 5.66 on Qwen2.5-7B,
7.16 to 8.76 on Qwen3-8B, and
10.66 to 11.88 on LLaVA-2-8B.
Its GPA gains over DPO are 1.16, 0.41, and
1.86 points, respectively.
Improvements also extend to
$\mathrm{GPA}_{\mathrm{clip}}$ and
$\mathrm{TSA}_{\mathrm{w}}$ on all three backbones,
indicating gains in both credited action content
and consecutive-action coverage.

\textbf{Effect of Severity-Dependent Weighting.}
Weighting alone increases GPA by 0.25, 0.27, and
1.08 points on the three backbones, but its effects
on other metrics are mixed.
On Qwen2.5-7B, $\mathrm{TSA}_{\mathrm{w}}$ decreases
from 17.01 to 16.52; on LLaVA-2-8B,
$\mathrm{GPA}_{\mathrm{clip}}$ declines from
29.15 to 28.97 despite the higher GPA.
Thus, weighting alone improves penalty-aware
performance without consistently increasing
credited content or sequence coverage.

\textbf{Combining Both Components.}
The full objective achieves the highest GPA, $\mathrm{GPA}_{\mathrm{clip}}$, and $\mathrm{TSA}_{\mathrm{w}}$ within every ablation group. On Qwen3-8B, joint optimization reaches 23.06 GPA, compared with 19.93 for weighting alone and 20.07 for the margin alone. On LLaVA-2-8B, it further improves $\mathrm{TSA}_{\mathrm{w}}$ from the margin-only result of 25.27 to 26.34. These results support combining both components, though joint gains vary across backbones. The ablation establishes the full formulation's benefit; isolating severity assignments from overall weighting and margin strength requires additional controls.

\begin{table*}[t]
\centering
\caption{Comprehensive ablation study of GM-DPO core components across all backbones on FlexBench. Omitting both components recovers the standard DPO baseline. Best performance within each model family is in \textbf{bold}.}
\label{tab:appendix_full_ablation}
\resizebox{0.96\linewidth}{!}{
\begin{tabular}{l|cc|cccc|cccc}
\toprule
\multicolumn{1}{c|}{\bf Model} & \bf Weight & \bf Margin & \multicolumn{4}{c|}{\bf Motion Accuracy (Penalty-Aware)} & \multicolumn{4}{c}{\bf Motion Sequence (Non-Penalty)} \\
& & & \multicolumn{1}{c}{\bf Subs.} & \multicolumn{1}{c}{\bf Non-subs.} & \multicolumn{1}{c}{\bf GPA} & \multicolumn{1}{c|}{$\bf GPA_{\text{clip}}$} & \multicolumn{1}{c}{$\bf Seq_2$} & \multicolumn{1}{c}{$\bf Seq_3$} & \multicolumn{1}{c}{$\bf Seq_{\text{chain}}$} & \multicolumn{1}{c}{$\bf TSA_{\text{w}}$} \\
\midrule
\multirow{4}{*}{Qwen2.5-7B} 
& \xmark & \xmark & 3.92  & 24.22 & 12.04 & 25.05 & 15.13 & 15.96 & 18.39 & 17.01 \\
& \cmark & \xmark & 4.19  & 24.44 & 12.29 & 25.00 & 15.86 & 14.61 & 17.92 & 16.52 \\
& \xmark & \cmark & 5.66  & 24.51 & 13.20 & 25.44 & 16.16 & 15.89 & 19.47 & 17.73 \\
& \cmark & \cmark & \textbf{6.14} & \textbf{25.95} & \textbf{14.06} & \textbf{25.82} & \textbf{17.05} & \textbf{17.31} & \textbf{19.72} & \textbf{18.46} \\
\midrule
\multirow{4}{*}{Qwen3-8B}   
& \xmark & \xmark & 7.16  & 38.40 & 19.66 & 28.86 & 17.69 & 19.94 & 20.52 & 19.78 \\
& \cmark & \xmark & 8.19  & 37.53 & 19.93 & 29.42 & 19.25 & 18.98 & 20.67 & 19.88 \\
& \xmark & \cmark & 8.76  & 37.04 & 20.07 & 29.82 & 19.40 & 19.18 & 20.79 & 20.03 \\
& \cmark & \cmark & \textbf{9.30} & \textbf{43.69} & \textbf{23.06} & \textbf{30.52} & \textbf{21.81} & \textbf{20.58} & \textbf{23.60} & \textbf{22.34} \\
\midrule
\multirow{4}{*}{LLaVA-2-8B} 
& \xmark & \xmark & 10.66 & 40.73 & 22.69 & 29.15 & 21.17 & 22.53 & 23.33 & 22.66 \\
& \cmark & \xmark & 10.23 & \textbf{44.09} & 23.77 & 28.97 & 21.52 & 23.66 & 23.73 & 23.27 \\
& \xmark & \cmark & 11.88 & 43.55 & 24.55 & 30.63 & 24.81 & 25.00 & 25.62 & 25.27 \\
& \cmark & \cmark & \textbf{13.89} & 42.66 & \textbf{25.40} & \textbf{31.77} & \textbf{25.87} & \textbf{25.49} & \textbf{27.04} & \textbf{26.34} \\
\bottomrule
\end{tabular}
}
\end{table*}

\subsection{Extended Downstream Generalization Benchmarks}
\label{sec:appendix_generalization}

To evaluate whether preference optimization on fine-grained physical dynamics induces catastrophic forgetting or impairs general multimodal reasoning, we benchmark aligned policies across three established downstream suites: Video-MME~\citep{fu2024videomme} for duration-stratified video understanding, POPE~\citep{li2023evaluating} for binary object hallucination diagnosis, and Video-ChatGPT~\citep{maaz2024videochatgpt} for open-ended conversational generation.

\textbf{Comprehensive Temporal Reasoning on Video-MME.}
As shown in Table~\ref{tab:appendix_videomme},
GM-DPO achieves overall accuracies of 60.62\%,
62.72\%, and 65.39\%, improving over the corresponding
base models by 1.25, 1.02, and 0.69 percentage points.
It also exceeds DPO on all three duration splits
for every backbone.
The gains vary by duration: on LLaVA-2-8B,
the improvement is larger on Short videos
(1.11 percentage points) than on Long videos
(0.60 percentage points).
These results support retained performance across
video durations following motion-focused alignment.

\begin{table*}[t]
\centering
\caption{Generalization performance on Video-MME across video duration splits and overall scores. Reference base models are shown in \textit{\textcolor{gray}{gray italics}}; within each backbone family, the best result among aligned models is in \textbf{bold}, and the second-best is \underline{underlined}. The overall column is shaded in \colorbox{highlightblue}{blue}.}
\label{tab:appendix_videomme}
\resizebox{0.85\linewidth}{!}{
\setlength{\tabcolsep}{18pt}
\begin{tabular}{l|ccc|>{\columncolor{highlightblue}}c}
\toprule
\textbf{Model Variant} & \textbf{Short} & \textbf{Medium} & \textbf{Long} & \textbf{Overall} \\
\midrule
\textcolor{gray}{\textit{Qwen2.5-7B-Base}} & \textcolor{gray}{\textit{68.89}} & \textcolor{gray}{\textit{59.11}} & \textcolor{gray}{\textit{50.11}} & \textcolor{gray}{\textit{59.37}} \\
Qwen2.5-7B-DPO   & 69.11 & \textbf{60.18} & 50.89 & 60.06 \\
Qwen2.5-7B-cDPO  & 68.89 & 59.53 & 50.33 & 59.58 \\
Qwen2.5-7B-IPO   & \underline{69.56} & \textbf{60.18} & \underline{51.61} & \underline{60.45} \\
Qwen2.5-7B-SimPO & 69.40    & 60.05    & 51.34    & 60.26    \\
Qwen2.5-7B-GM-DPO (Ours) & \textbf{69.75} & \underline{60.17} & \textbf{51.93} & \textbf{60.62} \\
\midrule
\textcolor{gray}{\textit{Qwen3-8B-Base}} & \textcolor{gray}{\textit{73.00}} & \textcolor{gray}{\textit{59.21}} & \textcolor{gray}{\textit{52.88}} & \textcolor{gray}{\textit{61.70}} \\
Qwen3-8B-DPO     & 73.33 & 59.60 & 54.11 & 62.35 \\
Qwen3-8B-cDPO    & 73.89    & \textbf{59.62}    & \textbf{54.67}    & \textbf{62.73}    \\
Qwen3-8B-IPO     & \underline{74.00} & 58.78 & 53.44 & 62.07 \\
Qwen3-8B-SimPO   & 73.67 & 59.06 & 54.00 & 62.24 \\
Qwen3-8B-GM-DPO (Ours)   & \textbf{74.22} & \underline{59.61} & \underline{54.34} & \underline{62.72} \\
\midrule
\textcolor{gray}{\textit{LLaVA-2-8B-Base}} & \textcolor{gray}{\textit{77.56}} & \textcolor{gray}{\textit{61.44}} & \textcolor{gray}{\textit{55.11}} & \textcolor{gray}{\textit{64.70}} \\
LLaVA-2-8B-DPO   & 77.56 & 61.22 & 55.22 & 64.67 \\
LLaVA-2-8B-cDPO  & 78.50    & 61.30    & 55.62    & 65.14    \\
LLaVA-2-8B-IPO   & \underline{78.56} & \underline{61.33} & \underline{55.67} & \underline{65.19} \\
LLaVA-2-8B-SimPO & 78.22 & 60.56 & 55.56 & 64.78 \\
LLaVA-2-8B-GM-DPO (Ours) & \textbf{78.67} & \textbf{61.68} & \textbf{55.82} & \textbf{65.39} \\
\bottomrule
\end{tabular}
}
\end{table*}

\textbf{Cross-Domain Object Hallucination Suppression on POPE.}
On the POPE benchmark (Table~\ref{tab:appendix_pope}), GM-DPO demonstrates robust zero-shot transfer in suppressing existential object hallucinations across Adversarial, Popular, and Random subsets. GM-DPO attains the highest Overall Acc across all backbones ($87.56$ on Qwen2.5-7B, $89.40$ on Qwen3-8B, and $88.97$ on LLaVA-2-8B) alongside top balanced F1 scores ($86.15$, $89.02$, and $88.41$, respectively). Concurrently, baseline objectives such as IPO and SimPO exhibit mild degradation on LLaVA-2-8B (F1 falling to $88.11$ and $88.04$), suggesting that uncalibrated quadratic margins or rigid sequence-length penalties can skew response calibration under binary probes. In contrast, GM-DPO stabilizes Yes-ratio distributions around $40.12$--$46.53\%$, preventing the binary over-affirmation collapse common to multimodal alignment.

\begin{table*}[t]
\centering
\caption{Zero-shot object hallucination evaluation on POPE across adversarial, popular, and random splits. Best aligned results within each model group are highlighted in \textbf{bold}, and second-best are \underline{underlined}; \textbf{Overall Acc} is shaded in \colorbox{highlightblue}{light blue}, and \textbf{F1} is shaded in \colorbox{highlightpink}{light pink}.}
\label{tab:appendix_pope}
\resizebox{0.82\linewidth}{!}{
\begin{tabular}{l|ccc>{\columncolor{highlightblue}}c|c>{\columncolor{highlightpink}}c}
\toprule
\textbf{Model Variant} & \textbf{Adversarial} & \textbf{Popular} & \textbf{Random} & \textbf{Overall Acc} & \textbf{Yes\%} & \textbf{F1} \\
\midrule
\textcolor{gray}{\textit{Qwen2.5-7B-Base}} & \textcolor{gray}{\textit{86.50}} & \textcolor{gray}{\textit{87.37}} & \textcolor{gray}{\textit{88.23}} & \textcolor{gray}{\textit{87.37}} & \textcolor{gray}{\textit{40.03}} & \textcolor{gray}{\textit{85.97}} \\
Qwen2.5-7B-DPO   & 86.53          & 87.43          & 88.33 & 87.43          & \textbf{40.14} & \underline{86.06} \\
Qwen2.5-7B-cDPO  & \underline{86.57} & \underline{87.47} & 88.33 & \underline{87.46} & \underline{40.12} & 86.06 \\
Qwen2.5-7B-IPO   & 86.53          & 87.40          & \underline{88.40}          & 87.44          & 40.29          & \underline{86.09}          \\
Qwen2.5-7B-SimPO & 86.50          & 87.38          & 88.36          & 87.41          & 40.10          & 86.02          \\
Qwen2.5-7B-GM-DPO (Ours) & \textbf{86.67} & \textbf{87.53} & \textbf{88.47} & \textbf{87.56} & \underline{40.12} & \textbf{86.15} \\
\midrule
\textcolor{gray}{\textit{Qwen3-8B-Base}} & \textcolor{gray}{\textit{87.17}} & \textcolor{gray}{\textit{89.17}} & \textcolor{gray}{\textit{91.57}} & \textcolor{gray}{\textit{89.30}} & \textcolor{gray}{\textit{46.08}} & \textcolor{gray}{\textit{88.86}} \\
Qwen3-8B-DPO     & 87.00          & \textbf{89.30} & 91.80          & \underline{89.37} & 46.50          & \underline{88.98} \\
Qwen3-8B-cDPO    & \underline{87.02} & \underline{89.28}          & \underline{91.81} & \underline{89.37} & \underline{46.51} & \underline{88.98} \\
Qwen3-8B-IPO     & 87.01          & \underline{89.28}          & \underline{91.81} & \underline{89.37} & 46.48          & 88.96          \\
Qwen3-8B-SimPO   & 87.00          & 89.26          & 91.78          & 89.35          & 46.47          & 88.94          \\
Qwen3-8B-GM-DPO (Ours)   & \textbf{87.07} & \textbf{89.30} & \textbf{91.83} & \textbf{89.40} & \textbf{46.53} & \textbf{89.02} \\
\midrule
\textcolor{gray}{\textit{LLaVA-2-8B-Base}} & \textcolor{gray}{\textit{86.51}} & \textcolor{gray}{\textit{88.23}} & \textcolor{gray}{\textit{90.77}} & \textcolor{gray}{\textit{88.50}} & \textcolor{gray}{\textit{44.27}} & \textcolor{gray}{\textit{88.10}} \\
LLaVA-2-8B-DPO   & 86.72          & 88.50          & 91.13          & 88.78          & \textbf{45.20} & 88.31          \\
LLaVA-2-8B-cDPO  & \textbf{86.88} & \underline{88.61} & 91.16          & \underline{88.88} & 45.11 & \underline{88.34} \\
LLaVA-2-8B-IPO   & 86.67          & 88.37          & \underline{91.28} & 88.77          & 44.98          & 88.11          \\
LLaVA-2-8B-SimPO & 86.44          & 88.27          & 90.77          & 88.49          & 44.48          & 88.04          \\
LLaVA-2-8B-GM-DPO (Ours) & \underline{86.80} & \textbf{88.65} & \textbf{91.47} & \textbf{88.97} & \underline{45.16}          & \textbf{88.41} \\
\bottomrule
\end{tabular}
}
\end{table*}

\textbf{Open-Ended Video Question Answering on Video-ChatGPT.}
Table~\ref{tab:appendix_videochatgpt} details per-dimension evaluations on Video-ChatGPT across Generic Understanding, Temporal Reasoning, and Consistency suites. 
GM-DPO achieves the highest Overall score among the aligned variants on all three backbones: 2.51 on Qwen2.5-7B, 2.75 on Qwen3-8B, and 2.57 on LLaVA-2-8B. These correspond to improvements of 0.01, 0.05, and 0.06 over DPO, respectively. GM-DPO also leads Consistency on all three backbones, with scores of 2.99, 3.15, and 2.89, and achieves the best or tied-best Temporal Reasoning scores. The advantages are not uniform across every dimension: DPO scores higher on Qwen2.5-7B's Context dimension (2.64 versus 2.63), while cDPO scores higher on Qwen3-8B's Correctness dimension (2.57 versus 2.56). These results support competitive overall response quality, with consistent advantages in temporal reasoning and consistency among the compared alignment methods.

\begin{table*}[t]
\centering
\caption{Video-ChatGPT evaluation across five dimensions using Qwen-3.8-Max as the judge on a 1--5 scale. Overall is the unweighted mean of the five scores, rounded to two decimal places. Base models are shown in \textit{\textcolor{gray}{gray italics}}. Best and second-best aligned results within each backbone are \textbf{bold} and \underline{underlined}, respectively; ties at the displayed precision share the same formatting.}
\label{tab:appendix_videochatgpt}
\resizebox{0.85\linewidth}{!}{
\begin{tabular}{l|ccc|c|c|>{\columncolor{highlightblue}}c}
\toprule
& \multicolumn{3}{c|}{\textbf{Generic Understanding}} & \textbf{Temporal} & \textbf{Consistency} & \textbf{Overall} \\
\textbf{Model Variant} & \textbf{Correctness} & \textbf{Detail} & \textbf{Context} & \textbf{Reasoning} & & \\
\midrule
\textcolor{gray}{\textit{Qwen2.5-7B-Base}}
& \textcolor{gray}{\textit{2.33}}
& \textcolor{gray}{\textit{2.23}}
& \textcolor{gray}{\textit{2.63}}
& \textcolor{gray}{\textit{2.16}}
& \textcolor{gray}{\textit{2.91}}
& \textcolor{gray}{\textit{2.45}} \\
Qwen2.5-7B-DPO
& \textbf{2.32} & \underline{2.33} & \textbf{2.64}
& 2.22 & \underline{2.98} & \underline{2.50} \\
Qwen2.5-7B-cDPO
& \underline{2.31} & \textbf{2.35} & \underline{2.63}
& \underline{2.23} & \underline{2.98} & \underline{2.50} \\
Qwen2.5-7B-IPO
& 2.30 & 2.29 & 2.58 & 2.19 & 2.92 & 2.46 \\
Qwen2.5-7B-SimPO
& \underline{2.31} & 2.32 & 2.60
& \underline{2.23} & 2.96 & 2.48 \\
Qwen2.5-7B-GM-DPO (Ours)
& \textbf{2.32} & \textbf{2.35} & \underline{2.63}
& \textbf{2.25} & \textbf{2.99} & \textbf{2.51} \\
\midrule
\textcolor{gray}{\textit{Qwen3-8B-Base}}
& \textcolor{gray}{\textit{2.55}}
& \textcolor{gray}{\textit{2.57}}
& \textcolor{gray}{\textit{2.89}}
& \textcolor{gray}{\textit{2.52}}
& \textcolor{gray}{\textit{3.14}}
& \textcolor{gray}{\textit{2.73}} \\
Qwen3-8B-DPO
& 2.53 & 2.53 & 2.85 & 2.51 & 3.07 & 2.70 \\
Qwen3-8B-cDPO
& \textbf{2.57} & \underline{2.57} & \underline{2.90}
& \underline{2.53} & \underline{3.10} & \underline{2.73} \\
Qwen3-8B-IPO
& 2.47 & 2.49 & 2.79 & 2.48 & 3.07 & 2.66 \\
Qwen3-8B-SimPO
& 2.45 & 2.48 & 2.79 & 2.45 & 3.04 & 2.64 \\
Qwen3-8B-GM-DPO (Ours)
& \underline{2.56} & \textbf{2.58} & \textbf{2.91}
& \textbf{2.55} & \textbf{3.15} & \textbf{2.75} \\
\midrule
\textcolor{gray}{\textit{LLaVA-2-8B-Base}}
& \textcolor{gray}{\textit{2.27}}
& \textcolor{gray}{\textit{2.35}}
& \textcolor{gray}{\textit{2.56}}
& \textcolor{gray}{\textit{2.36}}
& \textcolor{gray}{\textit{2.81}}
& \textcolor{gray}{\textit{2.47}} \\
LLaVA-2-8B-DPO
& 2.31 & 2.45 & 2.56 & \underline{2.41} & 2.84 & 2.51 \\
LLaVA-2-8B-cDPO
& 2.33 & \underline{2.54} & \textbf{2.65}
& 2.40 & 2.87 & \underline{2.56} \\
LLaVA-2-8B-IPO
& 2.33 & 2.53 & \underline{2.64}
& 2.40 & \underline{2.88} & \underline{2.56} \\
LLaVA-2-8B-SimPO
& \underline{2.35} & 2.51 & 2.63
& \textbf{2.42} & 2.85 & 2.55 \\
LLaVA-2-8B-GM-DPO (Ours)
& \textbf{2.36} & \textbf{2.55} & \underline{2.64}
& \textbf{2.42} & \textbf{2.89} & \textbf{2.57} \\
\bottomrule
\end{tabular}
}
\end{table*}

\subsection{Qualitative Visualizations and Case Studies}
\label{sec:appendix_qualitative}

\begin{figure*}[h]
\centering
\includegraphics[width=0.98\linewidth]{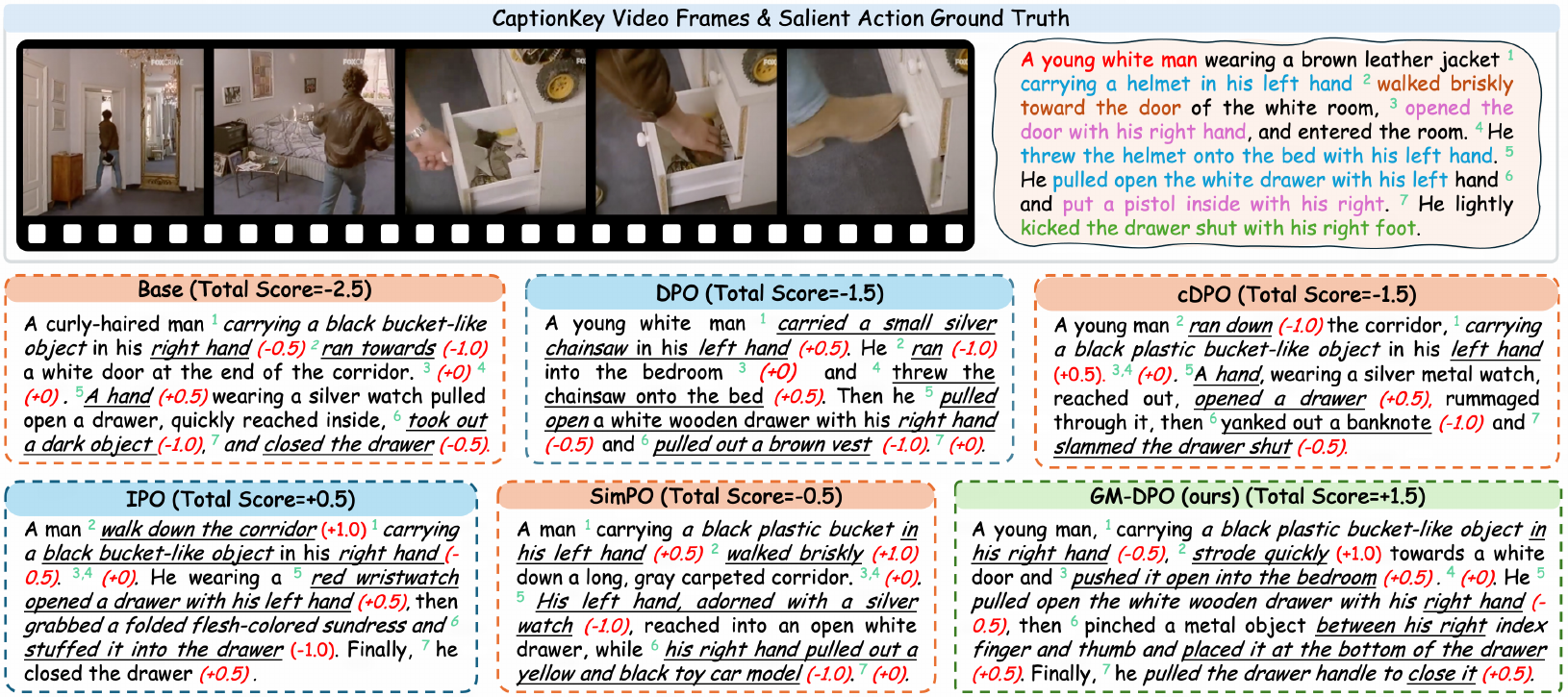}
\caption{Qualitative comparison of physical motion understanding on a representative multi-shot sequence from FlexBench. Predictions highlight salient limb manipulation and chronological interactions across 7 atomic checkpoints, with local premise scores annotated in parentheses ($-1.0$ for confabulations, $-0.5$ for chirality/imprecise errors, $+0.5/+1.0$ for grounded dynamics). Each block header reports the unnormalized raw cumulative score ($\text{Total Score} = \sum_{i=1}^7 s_i$).}
\label{fig:qualitative_comparison}
\end{figure*}

Figure~\ref{fig:qualitative_comparison} illustrates both
the relative improvement and remaining limitations
of GM-DPO.
It obtains the highest cumulative score in this
example (+1.5), compared with -2.5 for the base
model and -1.5 for DPO.
However, its caption still contains two laterality
errors, each assigned -0.5.
The example therefore illustrates improved aggregate
action fidelity rather than error-free limb grounding.

\section{Limitations and Future Work}
\label{sec:limitations}

Our results support incorporating action-error severity into preference learning to improve limb-motion fidelity. Despite consistent gains across three VLM backbones, several limitations warrant further investigation.

\paragraph{Extending the Grading Rubric to Broader Motion Domains.}
Our five-level action-grading rubric distinguishes correct and partially specified content, omissions, limb errors, and fabricated actions, yielding four perturbation-severity levels for training. It targets identifiable limb actions, including bimanual coordination and laterality, with explicit human-annotated reference facts. Extending this rubric to continuous, multi-phase activities requires additional criteria for whole-body transitions, movement speed, and deformable object interactions. Such activities may involve overlapping errors that a single perturbation category cannot capture. Future work could investigate richer grading schemes and their agreement with human judgments while preserving the distinction between incomplete coverage and incorrect assertions.

\paragraph{Ambiguity in Monocular Video.}
FlexBench and our training pipeline use monocular videos spanning diverse scenes and viewpoints. These inputs lack explicit depth measurements, while self-occlusion, extreme viewing angles, and camera cuts can obscure limb identity and motion. Such ambiguities may contribute to residual errors, although our experiments do not isolate their effects from limitations in model perception or caption generation. Skeletal representations or multi-view geometric priors may help resolve ambiguous interactions. Evaluating these approaches would require distinguishing recoverable information from details that remain unobservable.

\paragraph{Scope of Dense Caption Evaluation.}
Our evaluation focuses on limb-motion fidelity, per-person coverage, and shot structure. Caption length and structural compliance characterize the output but do not establish the accuracy or completeness of other content. Although experiments on three additional benchmarks provide evidence of broader multimodal performance, they do not directly assess camera angles, camera motion, scene details, or sound descriptions within dense captions. Dedicated evaluation of these dimensions, including audio-grounded assessment when audio inputs are available, remains future work.

\end{document}